# A novel approach to navigate the taxonomic hierarchy to address the Open-World Scenarios in Medicinal Plant Classification

Soumen Sinha* Tanisha Rana† Susmita Ghosh‡ Rahul Roy§

**Abstract**

In this article, we propose a novel approach for plant hierarchical taxonomy classification by posing the problem as an open class problem. It is observed that existing methods for medicinal plant classification often fail to perform hierarchical classification and accurately identifying unknown species, limiting their effectiveness in comprehensive plant taxonomy classification. Thus we address the problem of unknown species classification by assigning it best hierarchical labels. We propose a novel method, which integrates DenseNet121, Multi-Scale Self-Attention (MSSA) and cascaded classifiers for hierarchical classification. The approach systematically categorizes medicinal plants at multiple taxonomic levels, from phylum to species, ensuring detailed and precise classification. Using multi scale space attention, the model captures both local and global contextual information from the images, improving the distinction between similar species and the identification of new ones. It uses attention scores to focus on important features across multiple scales. The proposed method provides a solution for hierarchical classification, showcasing superior performance in identifying both known and unknown species. The model was tested on two state-of-art datasets with and without background artifacts and so that it can be deployed to tackle real word application. We used unknown species for testing our model. For unknown species the model achieved an average accuracy of 83.36%, 78.30%, 60.34% and 43.32% for predicting correct phylum, class, order and family respectively. Our proposed model size is almost four times less than the existing state of the art methods making it easily deploy able in real world application.

*Department of Computer Science and Engineering, Mahindra University, Bahadurpally, Hyderabad - 500043, India. (soumen20ucse179@mahindrauniversity.edu.in)

†Department of Computer Science and Engineering, Mahindra University, Bahadurpally, Hyderabad - 500043, India. (tanisha20ucse202@mahindrauniversity.edu.in)

‡Department of Computer Science and Engineering, Jadavpur University, Kolkata, West Bengal - 500043, India. (susmitaghoshju@jadavpuruniversity.edu.in)

§Corresponding author: Department of Computer Science and Engineering, Mahindra University, Bahadurpally, Hyderabad - 500043, India. (rahul.roy@mahindrauniversity.edu.in)

# 1 Introduction

Medicinal plant classification in the fields of botany, agriculture, and pharmacology is essential for identifying and utilizing plants for therapeutic purposes. Accurate classification allows researchers to systematically catalog and study plant species, which is crucial for discovering new medicinal properties and developing herbal remedies. Moreover, it helps in the conservation of biodiversity, ensuring that valuable plant species are protected and sustainably used. Furthermore, understanding the taxonomy of medicinal plants aids in tracing the evolutionary relationships providing insights into their potential health benefits.

This knowledge is vital for pharmacologists in formulating effective and safe herbal medicines, contributing to the advancement of alternative and complementary therapies.

The recent advancements in deep learning have significantly improved the methods for detecting and classifying plant diseases, thereby enhancing the precision and efficiency of these tasks. Li et al. [20] reviewed various deep learning models for plant disease detection, highlighting their potential to automate agricultural processes and increase accuracy. Similarly, Chen et al. [4] provided a comprehensive review on plant image recognition using deep learning, emphasizing the robustness of these models in handling diverse plant datasets. Fitzgerald et al. [8] discussed the historical and regional advancements in medicinal plant analysis, focusing on emergent complex techniques used in modern pharmacology. Diwedi et al. [7] proposed an classification system based on optimized support vector machine.

Tan et al. [21] explored the use of deep learning for plant species classification using leaf vein morphometrics, achieving notable improvements in classification accuracy. However, their method does not extend to the hierarchical classification. Ganzera and Sturm [9] highlighted recent advancements in HPLC/MS in medicinal plant analysis, focusing primarily on chemical analysis. Additionally, Vishnoi et al. [24] provided a comprehensive study of feature extraction techniques for plant leaf disease detection, emphasizing the importance of feature extraction in classification tasks for plants affected with various diseases.

Recent studies have applied various deep learning techniques for medicinal plant classification. Azadnia et al. [2] proposed a SCAM-herb based model which uses gated pooling methods to classify medicinal and poisonous plants from visual characteristics of leaves, demonstrating the potential of deep learning in this domain. Samuel et al. [17] focused on the antioxidant and phytochemical classification of medicinal plants used in cancer treatment where classification was based on the predominant antioxidant. Tan et al. [20] employed visual feature-based deep learning for rapid identification of medicinal plants, showing significant improvements in speed and accuracy.

Various researchers also explored the use of pixel-wise and constrained feature extraction to improve classification. Dhakal and Shakya [6] explored the use of pixel wise operations in image-based plant disease detection. Kan et al. [12] examined multi feature extraction techniques for plant leaf image classification. Sachar and Kumar [16] surveyed various feature extraction and classification techniques for identifying plants through leaves. Tiwari et al. [23] developed an interesting deep neural network for multi-class classification of medicinal plant leaves where they classified 12 distinct crops in 22 different categories. Several intra class and inter class variations were considered during training. Naresh et al. [14] proposed a modified Local binary patterns (MLBP) method extract texture features from plant leaves. This helped them to tackle the issue of capturing the texture of plant leaves belonging to same plant species. Dey et al. [5] assessed various deep convolutional neural network models for automated medicinal plant identification ([1], [13], [4], [18]) from leaf images and gave a comparative analysis on them.

The primary shortcoming of these existing methods were their inability to perform hierarchical classification and tackle the challenge of identifying unknown or new species, which is crucial for a comprehensive medicinal plant classification. Many current approaches focus on known species and do not extend to the hierarchical classification needed for effectively dealing with new species. Hierarchical classification is essential as it allows for the systematic categorization of plants at various taxonomic levels, from phylum down to species. This level of detail is particularly important for accurately distinguishing between closely related species and understanding their evolutionary relationships. Without hierarchical classification, the ability to fully utilize the vast potential of medicinal plants is hindered, especially in discovering and categorizing new species with potential therapeutic benefits.

In this study, we propose a novel method , which integrates DenseNet121 , Multi-Scale Self-Attention (MSSA) and cascaded classifiers for hierarchical classification of medicinal plants. This approach not only used to classifying known species but also used to identify hierarchial classification for unknown species. By integrating MSSA, our model improves the feature representation technique by capturing both local and global contextual information, which is critical for distinguishing between similar species and identifying new ones. Unlike other models that do not tackle hierarchical classification, our proposed model systematically classifies medicinal plants at multiple taxonomic levels(Phylum to Species), ensuring a more detailed and accurate classification process.

Therefore the summary of our contributions are as follows:

- We introduce a new method that integrates DenseNet121, Multi-Scale Self-Attention (MSSA) and cascaded classifiers for hierarchical classification.
- Unlike other models that don't handle hierarchical classification, our model classifies medicinal plants at multiple taxonomic levels ranging from phylum to species, leading to more precise and accurate classifications.
- Our method not only deals with known medicinal plants but also provides hierarchical classification when introduced to unknown/new medicinal plants.
- The proposed model is almost four times smaller in size as compared to other state of the art methods and it achieves great results when dealing with datasets with various background artifacts which makes it easily deploy able in real-world applications making it practical for widespread use.
- The proposed model achieves promising results not only for known medicinal plant species but also for unknown plant species.

The article is organized into five sections. Following introduction we have preliminaries for our proposed approach. Section 3 describes the proposed method. In Section 4 and 5 we have the discussion on experimental setup and experimental results. In the end, Section 6 contains the conclusion.

# 2 Preliminaries

In this section we present a brief discussion on the preliminaries of the technologies used in our proposed method, this includes discussion on DenseNet121 and Multi-Scale Self-Attention.

## 2.1 DenseNet121

The feature extractor used in our proposed method is DenseNet21 [11]. The DenseNet21 (Densely Connected Convolutional Network) is an extension of the DenseNet architecture that consists of 21 layers. DenseNet is known for its dense connectivity pattern, where each layer is connected to every other layer in a feed-forward manner. This design helps the flow of information and gradients throughout the network for more efficient training and better performance. The architecture is shown in Figure 1.

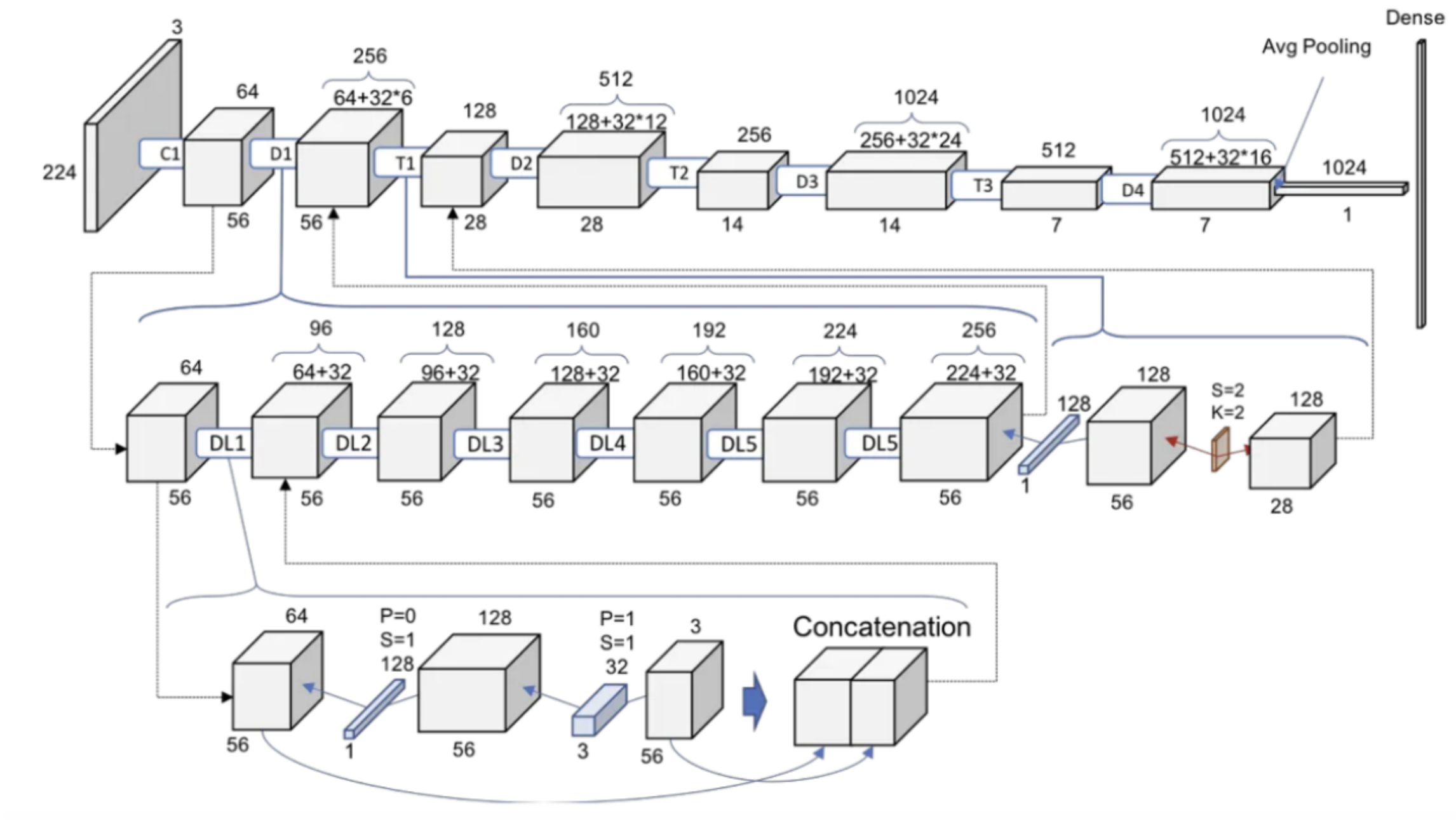

Figure 1: DenseNet121 feature extractor

DenseNet121 consists of four dense blocks interspersed with transition layers. Each dense block comprises several convolutional layers, with each layer receiving inputs from all preceding layers within the same block. This connectivity is mathematically repre-

sented as:

$$x_l = H_l([x_0, x_1, \ldots, x_{l-1}]) \tag{2.1}$$

where $x_l$ is the output of the $l$-th layer, $[x_0, x_1, \ldots, x_{l-1}]$ represents the concatenation of the feature maps produced by layers 0 to $l-1$, and $H_l(\cdot)$ is the composite function of the operations at the $l$-th layer, which includes Batch Normalization (BN), Rectified Linear Unit (ReLU), and Convolution (Conv). An illustration is shown in Figure 2.

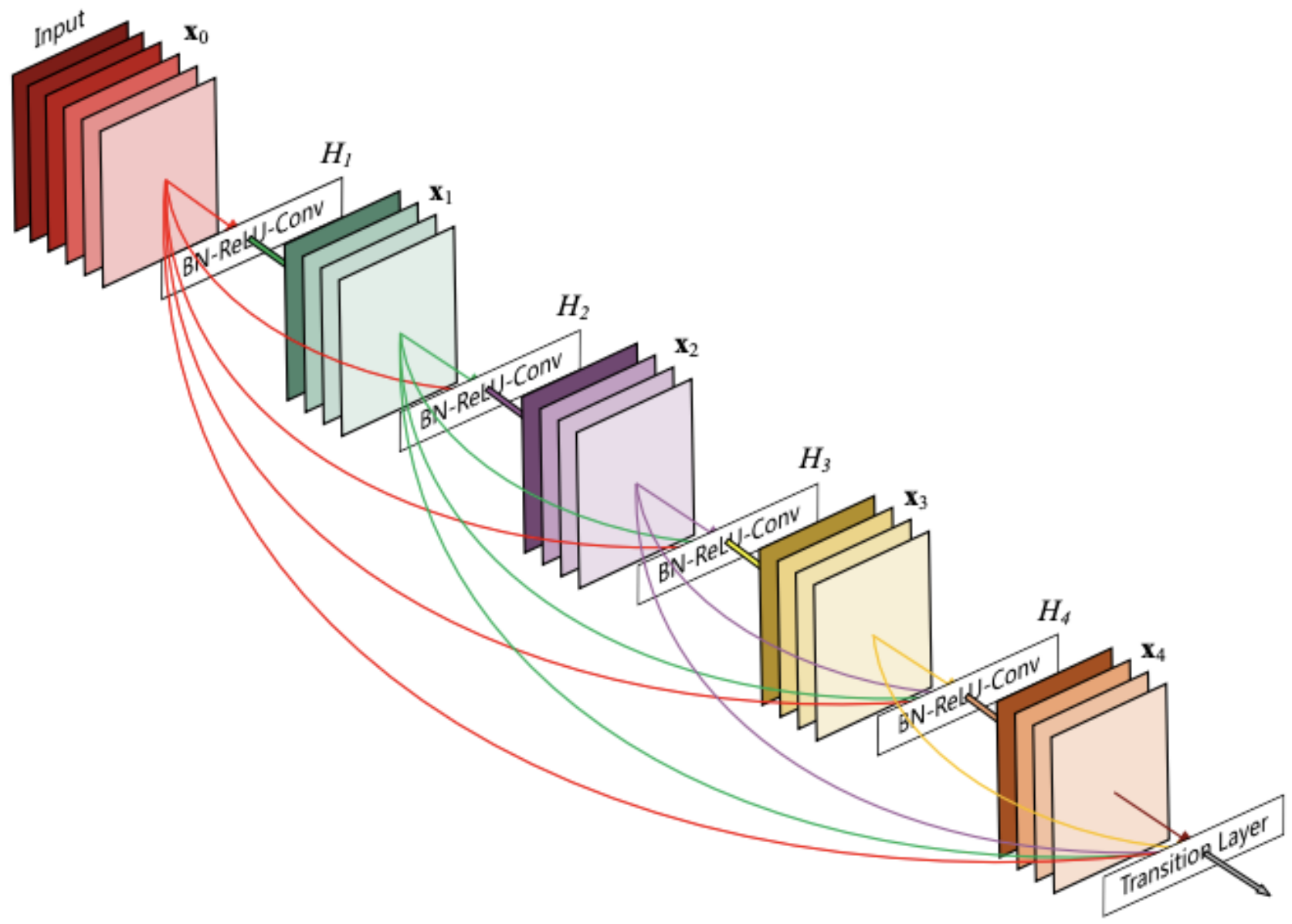

Figure 2: A 5-layer dense block with a growth rate of k = 4. Each layer takes all preceding feature-maps as input

The network begins with an initial convolution layer that processes the input image, followed by a pooling layer. The initial layers can be described as:

$$x_0 = \text{Conv}(\text{Input}), \quad x_1 = \text{Pooling}(x_0) \tag{2.2}$$

**Dense Block 1** consists of 6 convolutional layers, each densely connected to the previous layers. The transition layer following this block (2.3) includes batch normalization, a 1x1 convolutional layer, and a 2x2 average pooling layer. Similarly **Dense Block 2** consists of 12 convolutional layers, followed by another transition layer (2.4). **Dense Block 3** contains 24 convolutional layers. The transition layer here (2.5) also includes batch normalization, a 1x1 convolution, and a 2x2 average pooling. **Dense Block 4** is the final block with 16 convolutional layers, after which a global average pooling layer (2.6) reduces the spatial dimensions of the feature maps.

$$x_{\text{trans1}} = \text{AvgPool}(\text{Conv}_{1x1}(\text{BN}(x_1))) \tag{2.3}$$

$$x_{\text{trans2}} = \text{AvgPool}(\text{Conv}_{1x1}(\text{BN}(x_{\text{trans1}}))) \tag{2.4}$$

$$x_{\text{trans3}} = \text{AvgPool}(\text{Conv}_{1x1}(\text{BN}(x_{\text{trans2}}))) \tag{2.5}$$

$$x_{\text{gap}} = \text{GlobalAvgPool}(x_{\text{trans3}}) \tag{2.6}$$

Each layer in DenseNet121 typically consists of three operations:

$$H_l(x) = W_l * \sigma(BN(x)) \tag{2.7}$$

where $BN$ denotes Batch Normalization, $\sigma$ represents the ReLU activation function, $W_l$ is the weight matrix of the convolutional layer, and $*$ denotes the convolution operation. The growth rate $k$ is a crucial hyperparameter in DenseNet, indicating the number of feature maps added by each layer. If the input to the $l$-th layer has $m$ feature maps, the output will have $m + k$ feature maps. Thus, the width of the network grows linearly with the depth.

To control the complexity and size of the model, DenseNet121 employs transition layers between dense blocks, which consist of a Batch Normalization layer, a 1x1 Convolutional layer, and a 2x2 Average Pooling layer. The transition layer can be expressed as:

$$T(x) = \text{AvgPool}(W_t * \sigma(BN(x))) \tag{2.8}$$

where $W_t$ is the weight matrix of the 1x1 convolutional layer in the transition layer. In our proposed method we have used DenseNet121 as a feature extractor. Since each layer is connected to its previous layer DenseNet121 ensures efficient gradient propagation and better representation of the complex structural features of previous layers.

## 2.2 Multi-Scale Self-Attention (MSSA)

Self-attention techniques are widely used to compute contextual relationships and enhance the feature representation learned by convolutional layers. In self-attention, an input feature map is transformed into a weighted feature map that captures contextual relationships. However, this weighted feature map often lacks sufficient contextual information. Specifically, feature maps from shallow layers contain rich local spatial details but lack high-level semantics, while feature maps from deeper layers contain high-level semantic information but miss local spatial details.

To address these limitations, Multi-Scale Self-Attention (MSSA) [25] is used to integrate both local spatial and high-level semantic contextual information through multi-scale features learned by different convolutional blocks. The MSSA block takes a multi-scale feature map $F$ and a resized local feature map $C_i'$ as inputs, producing a weighted multi-scale feature map $D_i$ that captures contextual relationships among pixels from both local spatial and high-level semantic perspectives.

For example, consider five outputs from a feature extractor, each denoted as $C_i$, where $i$ ranges from 1 to 5, corresponding to different convolutional blocks. $C_i$ contains feature maps of varying scales at different depths, with scales decreasing and depth increasing as $i$ increases. To merge both local spatial details and high-level semantics, outputs from these five blocks are used to form a multi-scale feature map $F$. To retain local spatial details at the highest resolution, each output (e.g., $C_2, C_3, C_4$, and $C_5$) is resized to match the dimensions of $C_1$ by:

$$C'_i = \text{upsample}(C_i) \quad \text{and} \quad |C'_i| = |C_1| \tag{2.9}$$

where $i = 2, 3, 4$, and 5, and $|x|$ represents the dimension of a feature map $x$ without depth. All resized outputs are then concatenated to construct a multi-scale feature map $F$ by:

$$F = C'_1 \oplus C'_2 \oplus C'_3 \oplus C'_4 \oplus C'_5 \tag{2.10}$$

where $\oplus$ denotes the concatenation operation. Each high-resolution $C'_i$ and the multi-scale feature map $F$ are individually fed into the proposed MSSA module, which will be detailed in subsection 2.2, to compute contextual relationships.

For an input feature map $C'_i \in \mathbb{R}^{H\times W\times Ch_1}$, where $H$, $W$, and $Ch_1$ represent the height, width, and channel dimensions respectively, and $i$ denotes the block number, a 1x1 convolution is applied to transform $C'_i$ into a new feature map $Y \in \mathbb{R}^{H\times W\times \frac{Ch_1}{8}}$. A ratio of 1/8 is used to reduce the channel number to its 1/8, which has been empirically determined to be optimal [26]. Similarly, for the multi-scale feature map $F \in \mathbb{R}^{H\times W\times Ch_2}$, a 1x1 convolution is used to generate a new feature map $Z \in \mathbb{R}^{H\times W\times \frac{Ch_1}{8}}$.

We then reshape $Y$ to $Y_r$ of size $(H \times W) \times \frac{Ch_1}{8}$ and reshape and transpose $Z$ to $Z_{rt}$ of size $\frac{Ch_1}{8} \times (H \times W)$. Multiplying $Y_r$ and $Z_{rt}$ generates a map of size $(H \times W) \times (H \times W)$. Applying a softmax to this map produces a normalized map $A$, also known as the attention map. The attention map $A$ is computed as:

$$A(m,n) = \frac{\exp(Y_r(m,:) \cdot Z_{rt}(:,n))}{\sum_{n=1}^{H\times W} \exp(Y_r(m,:) \cdot Z_{rt}(:,n))} \tag{2.11}$$

where : denotes all values in a row or column, and $A(m,n)$ represents the impact of the $n$-th column of $Z_{rt}$ on the $m$-th row of $Y_r$. A high value in $A$ indicates a strong correlation between $Y_r$ and $Z_{rt}$ (i.e., between $C'_i$ and $F$).

In another branch, a 1x1 convolution transforms $C'_i$ into a new feature map $X \in \mathbb{R}^{H\times W\times Ch_1}$, which is then reshaped and transposed to $X_{rt}$ of size $Ch_1 \times (H \times W)$. A matrix multiplication between $X_{rt}$ and $A$ is performed, and the result is reshaped to size $H \times W \times Ch_1$ and scaled by a learnable parameter $\mu$ to generate a weighted attention map. This map is added to the input $C'_i$ to produce a weighted feature map $D_i$:

$$D_i(m,n) = \mu \cdot \text{reshape}(X_{rt}(m,:) \cdot A(:,n)) + C'_i(m,n) \tag{2.12}$$

where $D_i(m,n)$ represents the value of a weighted feature map at location $(m,n)$, and $\mu$ is initialized to 0 to allow the network to initially rely on local neighborhood cues to maximize learning.

Starting with $D_5$, a 3x3 filter is applied, and the filtered result is concatenated with $D_4$ to combine spatial and semantic information from blocks 5 and 4. This operation is repeated to combine information from blocks 4 and 3, blocks 3 and 2, and blocks 2 and 1. The algorithmic steps for chained concatenation operations are as follows:

**Algorithm 1** Chained Concatenation Operations in MSSA Module

Initialize $U_5 = D_5$
**for** $i = 5$ **to** $2$ **do**
  $U_{i-1} = \text{conv}(U_i) \oplus D_{i-1}$
**end for**

where $i$ represents the block number and $U_{i-1}$ contains spatial and semantic information from the $i$-th and $i-1$-th blocks. A 3x3 convolution is then applied to $U_1$, followed by bilinear interpolation and softmax to generate the segmentation result.

The final weighted multi-scale feature map $D_i$ thus integrates spatial details and semantic information, enhancing the feature representation for improved segmentation accuracy. Figure 3 gives an illustration of MSSA module.

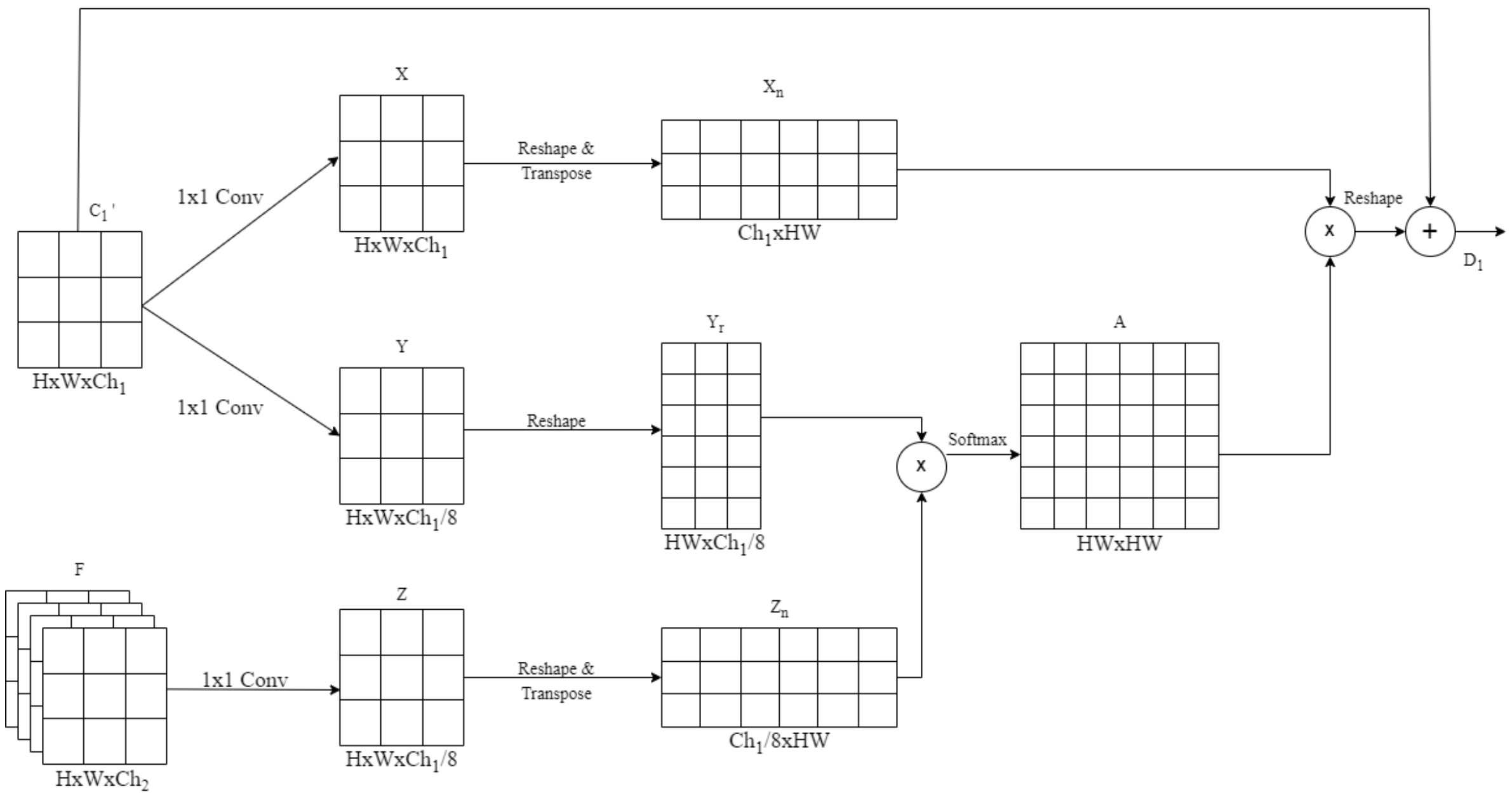

Figure 3: Illustration of MSSA module

# 3 Proposed Method

In this section, we propose a novel architecture (DenseNet121 with Multi-Scale Self-Attention and Cascaded Classifiers) designed for the hierarchical classification of medicinal plants and address the open world challenges. Our model uses DenseNet121 backbone as a feature extractor to extract the features. We integrate a Multi-Scale Self-Attention mechanism to our proposed method which in turn helps the model to learn different contextual relationship and important features for hierarchical classification . The MSSA mechanism processes multi-scale feature maps from different convolutional layers of DenseNet121, allowing the model to capture both local spatial details and high-level semantic information. Subsequently, cascaded classifiers are employed to predict taxonomic categories (Phylum, Class, Order, Family, Genus, and Species) in a hierarchical manner.

## 3.1 Problem Description

Traditional classifiers used for medicinal plant classification are unable to perform hierarchical classification at the taxonomic level. This poses a severe challenge for these models as they fail to predict the taxonomic categories when an unknown or new species is discovered. To bridge this challenge we propose a novel architecture that gives us hierarchical classification of medicinal plants. Our proposed model also addresses the challenge of unknown/new species classification. In the following section we discuss in detail our proposed method.

## 3.2 Model Architecture

The proposed architecture is designed to effectively classify medicinal plants by leveraging rich feature representations and contextual relationships. The architecture integrates DenseNet121 as the backbone network for feature extraction, a Multi-Scale Self-Attention (MSSA) module to enhance contextual relationships, and a series of cascaded classifiers for hierarchical classification. The proposed model architecture is shown in Figure 4

### 3.2.1 DenseNet121 Backbone

DenseNet121 is used as the backbone network due to its dense connectivity, which promotes feature reuse and efficient gradient flow. The DenseNet121 architecture consists of multiple dense blocks, each containing several convolutional layers. The output of each layer is concatenated with the outputs of all preceding layers within the same block, allowing the network to learn robust feature representations. It acts as a feature extractor for our proposed method.

### 3.2.2 Multi-Scale Self-Attention (MSSA)

The MSSA module is used with DenseNet121 as its backbone to capture local spatial details and high-level semantic information by applying self-attention to multi-scale feature maps obtained from different convolutional blocks of DenseNet121. In our implementation, we utilize three key layers **('pool2', 'pool3','pool4')** from DenseNet121 for MSSA. These layers correspond to the outputs after the second, third, and fourth pooling layers, respectively. The 'pool2' layer has the highest resolution with dimensions $H_1 \times W_1 \times C_1$, while the 'pool3' and 'pool4' layers have progressively smaller resolutions and more channels, denoted as $H_2 \times W_2 \times C_2$ and $H_3 \times W_3 \times C_3$, respectively. For our MSSA module, we denote these feature maps as $F_1$, $F_2$, and $F_3$, respectively.

The MSSA mechanism can be described as follows:

Let $C_i$ denote the output of the $i$-th selected convolutional block of DenseNet121, where $i$ ranges from 1 to 3. To maintain high-resolution spatial details, we resize each

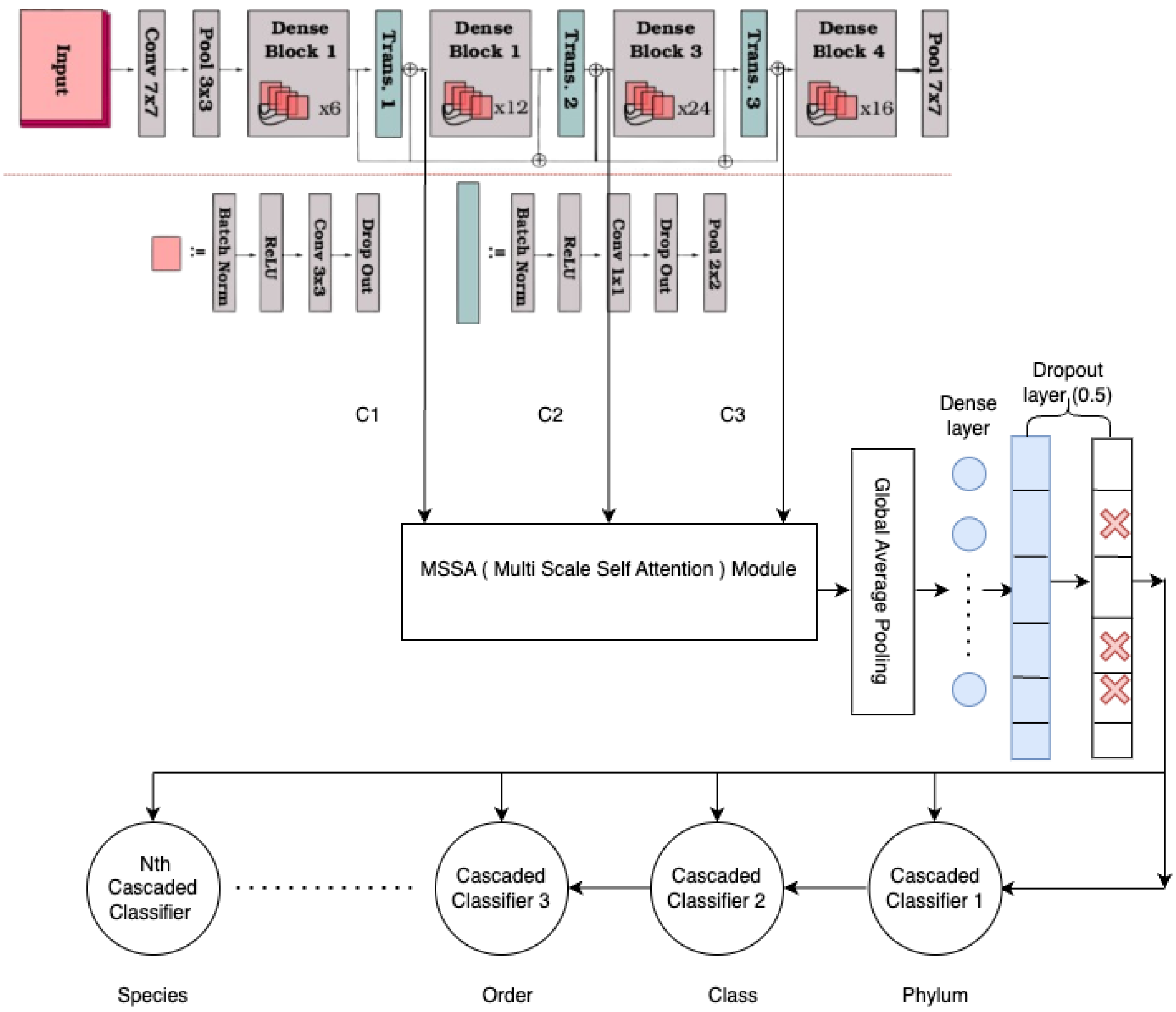

Figure 4: Proposed Model Architecture for hierarchical taxonomy generation for plants

output $C_i$ to match the dimension of the highest resolution output $C_1$:

$$C_i' = \text{resize}(C_i, \text{shape}(C_1)) \quad \text{and} \quad |C_i'| = |C_1| \tag{3.1}$$

Next, we concatenate the resized outputs to construct a multi-scale feature map $F$:

$$F = C_1' \oplus C_2' \oplus C_3' \tag{3.2}$$

where $\oplus$ denotes the concatenation operation. For each input feature map $C_i' \in \mathbb{R}^{H\times W\times Ch_1}$, where $H$, $W$, and $Ch_1$ represent the height, width, and channel dimensions respectively, we apply a $1\times 1$ convolution to transform $C_i'$ into a new feature map $Y \in \mathbb{R}^{H\times W\times \frac{Ch_1}{8}}$:

$$Y = \text{Conv}_{1\times 1}(C_i') \tag{3.3}$$

Similarly, for the multi-scale feature map $F \in \mathbb{R}^{H\times W\times Ch_2}$, we apply a 1x1 convolution to generate a new feature map $Z \in \mathbb{R}^{H\times W\times \frac{Ch_1}{8}}$:

$$Z = \text{Conv}_{1\times 1}(F) \tag{3.4}$$

We then reshape $Y$ to $Y_r$ of size $(H \times W) \times \frac{Ch_1}{8}$ and reshape and transpose $Z$ to $Z_{rt}$ of size $\frac{Ch_1}{8} \times (H \times W)$. The attention map $A$ is computed as 2.11

In a parallel branch, we apply another 1x1 convolution to transform $C_i'$ into a new feature map $X \in \mathbb{R}^{H \times W \times Ch_1}$, which is then reshaped and transposed to $X_{rt}$ of size $Ch_1 \times (H \times W)$.

The weighted attention map is computed as 2.12, where $D_i(m, n)$ is the value of a weighted feature map at location $(m, n)$, and $\mu$ is a learnable parameter initialized to 0. Starting with $D_3$, we apply a 3x3 convolution and concatenate the result with $D_2$ to integrate spatial and semantic information:

$$U_3 = D_3 \tag{3.5}$$

$$U_{i-1} = \text{conv}(U_i) \oplus D_{i-1} \quad \text{for} \quad i = 3 \text{ to } 2 \tag{3.6}$$

The algorithmic steps for chained concatenation operations are as follows:

**Algorithm 2** Chained Concatenation Operations in MSSA Module

1: Initialize $U_3 = D_3$
2: **for** $i = 3$ **to** 2 **do**
3: $\quad U_{i-1} = \text{conv}(U_i) \oplus D_{i-1}$
4: **end for**

Finally, we apply a 3×3 convolution to $U_1$, followed by bilinear interpolation and softmax to generate the attention based feature map. Bilinear interpolation method is used to resample image pixels, providing a smoother and more accurate output than nearest-neighbor interpolation. The new pixel value $P$ is computed as a weighted average of the four nearest pixel values. Let $Q_{11}$, $Q_{12}$, $Q_{21}$, and $Q_{22}$ be the four nearest pixels to the target pixel, where $Q_{11}$ is the top-left pixel, $Q_{12}$ is the top-right pixel, $Q_{21}$ is the bottom-left pixel, and $Q_{22}$ is the bottom-right pixel. The interpolated value $P$ is computed as follows:

$$P = Q_{11}(1 - x)(1 - y) + Q_{21}x(1 - y) + Q_{12}(1 - x)y + Q_{22}xy \tag{3.7}$$

where $x$ and $y$ are the relative distances of the target pixel from the top-left corner within the unit square.

### 3.2.3 Cascaded Classifiers

The output of the MSSA module is passed through a Global Average Pooling layer to reduce the spatial dimensions, followed by a dense layer with 512 units and ReLU activation, and a dropout layer with a dropout rate of 0.5. Subsequently, the model employs a series of cascaded classifiers to predict taxonomic categories (Phylum, Class, Order, Family, Genus, and Species) in a hierarchical manner. Each classifier branch consists of a dense layer with 256 units and ReLU activation, followed by a softmax output layer tailored to the number of classes in each category. The output of each classifier is concatenated with the input features and fed into the subsequent classifier, ensuring a hierarchical prediction structure. The architecture of each cascaded classifier includes a dense layer with 256 units and ReLU activation, which receives the input features. This dense layer is followed by a softmax output layer that outputs the probability distribution over the classes for the current taxonomic category. The output of the softmax layer is then concatenated with the input features to form the input for the next classifier branch. The use of cascaded classifiers enables us to build a hierarchical system and moreover it helps us to generalize to unknown species based on shared taxonomic characteristics.

## 3.3 Unknown Species Prediction

**Algorithm 3** Hierarchical Classification with Confidence Threshold

**Require:** Predicted class probabilities, Confidence Threshold
**Ensure:** Final Taxonomic Classification

```
Set the initial taxonomic category to Phylum
for each taxonomic level (Phylum, Class, Order, Family, Genus, Species) do
    if Predicted probability at the current level ≥ Confidence Threshold then
        Assign the sample to the predicted class at this level
        Move to the next taxonomic level
    else
        Stop classification and return classification results BREAK
    end if
end for
```

For unknown species classification, the model employs a hierarchical classification approach, where predictions are made at various taxonomic levels, including Phylum, Class, Order, Family, Genus, and Species. To determine class membership, a confidence threshold of 0.6 is applied. If the predicted class probability surpasses this threshold, the sample is assigned to that particular class. This threshold is set considering potential noise in the images. The hierarchical process proceeds from broader categories to more specific ones. However, if the confidence falls below 0.6 at any level, the classification process terminates, and the classification taxonomy is generated till that point and is returned as an output to us. This strategy helps ensure robust classification while accommodating the possibility of ambiguous or uncertain predictions. The hierarchical cascaded classifier, featuring an adaptive confidence threshold, proves invaluable for classifying previously unknown species not encountered during training. This approach effectively addresses the challenge of open-world recognition by allowing the system to make informed decisions when faced with unfamiliar species.

# 4 Experimental Setup

## 4.1 Dataset Description

In this study, we utilize two types of datasets to evaluate the performance of the proposed model: one dataset with background artifacts and another dataset without background artifacts. The dataset with noise (DIMPSAR), sourced from Kaggle [3], includes images of medicinal plants taken in real-world settings. These images contain various types of background artifacts, such as objects, and environmental factors. This dataset simulates realistic conditions where image data may include additional elements that can interfere with the clear identification of the medicinal plants. The presence of background artifacts in the images presents a significant challenge for the model, testing its robustness and ability to handle real and cluttered data effectively. The dataset without background artifacts, obtained from Mendeley Data [15], consists of clean images of medicinal plants, taken in controlled environments without any background noise. This dataset is used to assess the model's performance under ideal conditions, providing a baseline for comparison with the noisy dataset. The artifact-free images allow the model to focus on learning the intrinsic features of the plants without the interference of extraneous background elements.

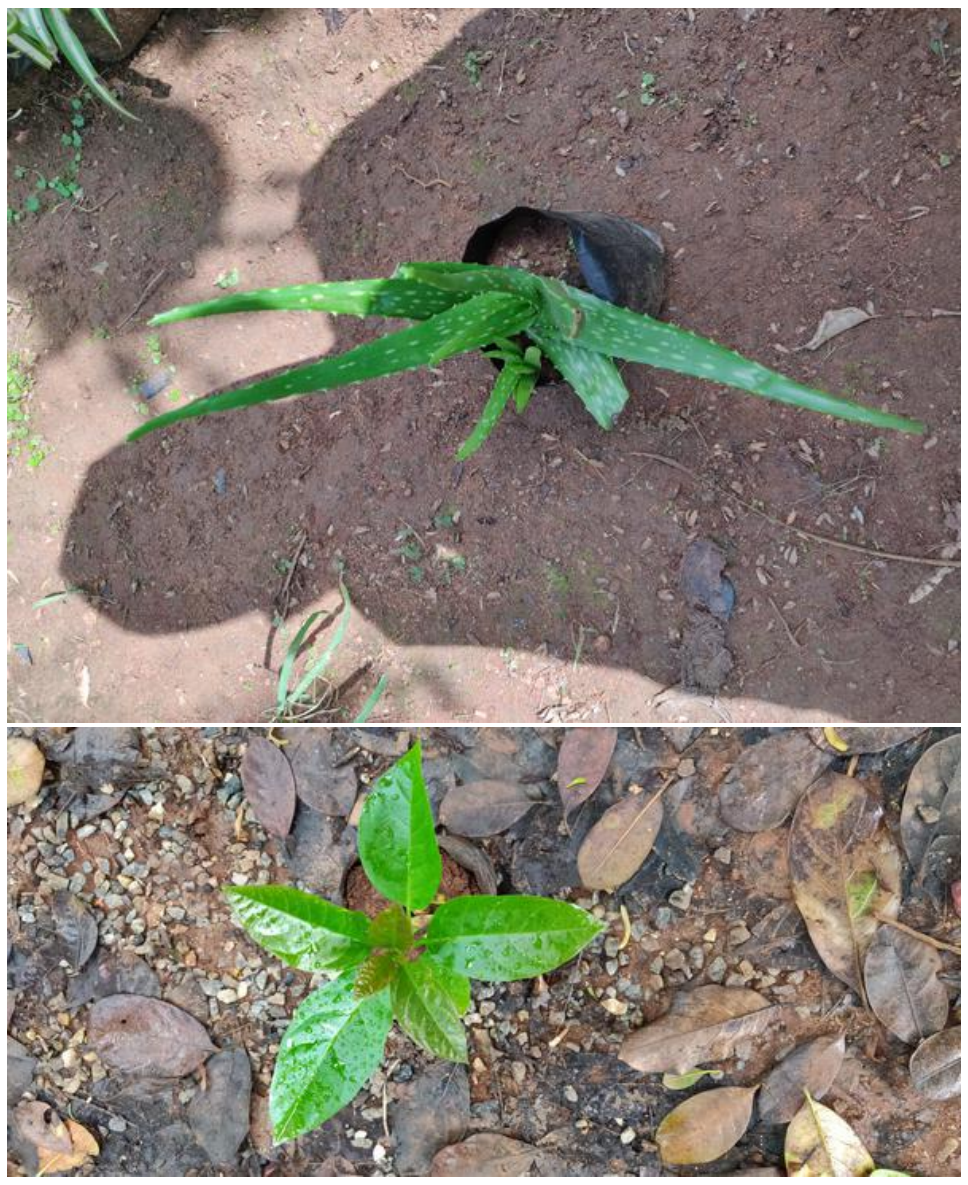

(a) Dataset with background artifacts

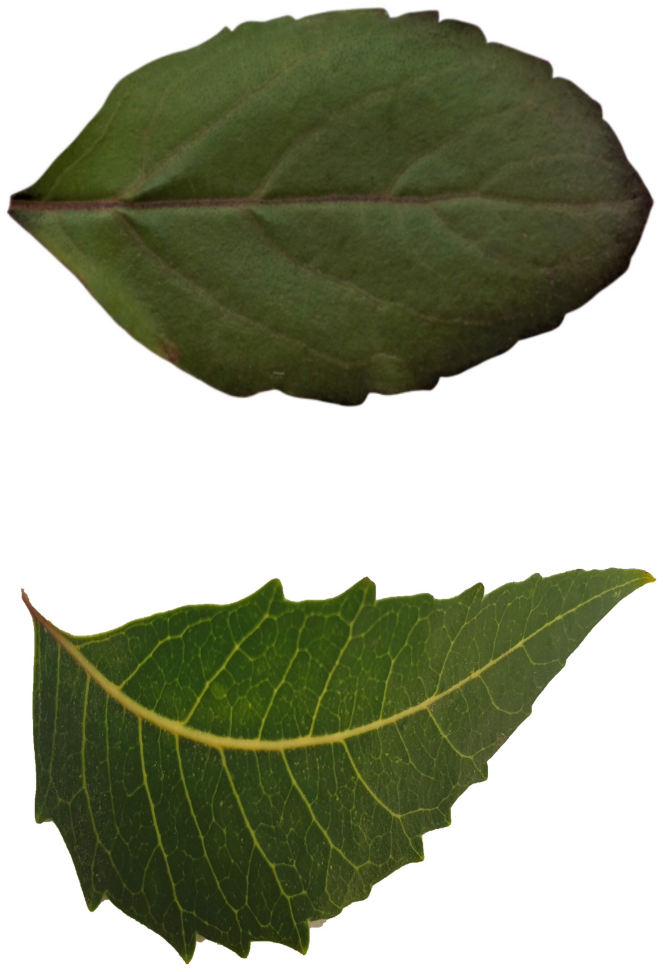

(b) Dataset without background artifacts

Figure 5: Dataset Visualisation

Figure 5 shows Noisy dataset consists of 10 unique medicinal plant species and non-noisy dataset consists of 15 unique medicinal plant species. Detailed explanation on dataset is given on Table 1 and 2. Noisy dataset has 10 species, 10 genus, 10 families, 10 orders, 3 classes and 4 phylums. Similarly dataset without noise has 14 species, 14 genus, 12 families, 11 orders , 2 classes and 3 phylums. By noisy dataset we mean the dataset where the background environment is still intact.

## 4.2 Experimental Environment

In this study, all the experiments were performed on an NVIDIA DGX-1 supercomputer. The supercomputer has the following configuration: Dual 20 Core Intel Xeon E5-2698 V4 clocked at 2.2 GHz, 5120 NVIDIA cores, 512 GB 2.133 GHz DDR4 RDIMM (RAM). All the codes are written in Python version 3.9.13. In the next section we discuss and compare the experimental results.

Table 1: Dataset(without background artifacts) Description

| **Phylum** | **Class** | **Order** | **Family** | **Genus** | **Species** |
|---|---|---|---|---|---|
| Spermatophyta | Magnoliopsida | Lamiales | Lamiaceae | Ocimum | Tulasi |
| Magnoliophyta | Magnoliopsida | Scrophulariales | Oleaceae | Jasminum | Jasmine |
| | | Brassicales | Moringaceae | Moringa | Drumstick |
| | | Santalales | Santalaceae | Santalum | Sandalwood |
| | | Lamiales | Lamiaceae | Lamiaceae | Mint |
| | | Piperales | Piperaceae | Piper | Betel |
| | | Rosales | Moraceae | Ficus | Peepal Tree |
| | | Sapindales | Meliaceae | Azadirachta | Neem |
| | | Gentianales | Anacardiaceae | Mangifera | Mango |
| | | | Fabaceae | Trigonella | Fenugreek |
| | | | Apocynaceae | Carissa | Karanda |
| Tracheophyta | Magnoliopsida | Lamiales | Lamiaceae | Plectranthus | Mexican Mint |
| | | Myrtales | Myrtaceae | Psidium | Guava |
| | | Malvales | Malvaceae | Hibiscus | Hibiscus rosa-sinensis |
| | Liliopsida | Zingiberales | Zingiberaceae | Alpinia | Rasna |

Table 2: Dataset (with background artifacts) Description

| **Phylum** | **Class** | **Order** | **Family** | **Genus** | **Species** |
|---|---|---|---|---|---|
| Spermatophyta | Magnoliopsida | Lamiales | Lamiaceae | Ocimum | Tulasi |
| | Angiospermae | Laurales | Lauraceae | Persea | Avocado |
| Magnoliophyta | Magnoliopsida | Piperales | Piperaceae | Piper | Betel |
| | | Sapindales | Meliaceae | Azadirachta | Neem |
| | | Solanales | Solanaceae | Withania | Ashwagandha |
| | Liliopsida | Cyperales | Poaceae Barnhart | Cymbopogon Spreng | Lemon grass |
| Tracheophyta | Magnoliopsida | Malpighiales | Phyllanthaceae | Phyllanthus | Amla |
| | | Myrtales | Myrtaceae | Psidium | Guava |
| | | Malvales | Malvaceae | Hibiscus | Hibiscus |
| Anthophyta | Liliopsida | Asparagales | Asparagaceae | Aloe | Aloevera |

# 5 Results Analysis

## 5.1 Model Comparison

The experiments were performed on two datasets, and the proposed approach was compared with 6 approaches using accuracy as a metric for evaluation. The results are depicted in Table 3 and 4 respectively.

Table 3: Comparison of Model Performances (**Dateset without background artifacts**)

| Model | Phylum Acc. | Class Acc. | Order Acc. | Family Acc. | Genus Acc. | Species Acc. |
|---|---|---|---|---|---|---|
| EfficientNet [22] | 78.63% | 93.89% | 16.79% | 16.79% | 10.69% | 10.69% |
| DenseNet [11] | 99.24% | 100.00% | 99.24% | 99.24% | 99.24% | 99.24% |
| ResNet [10] | 78.63% | 93.89% | 10.69% | 6.87% | 12.21% | 12.21% |
| VGG [19] | 96.24% | 94.31% | 93.20% | 93.13% | 91.24% | 90.24% |
| VGG-MSSA | 96.24% | 96.00% | 93.95% | 94.71% | 95.47% | 95.47% |
| ResNet50-MSSA | 78.44% | 95.3% | 48.67% | 44.85% | 49.44% | 50.96% |
| **Proposed Method** | **99.24%** | **99.24%** | **99.24%** | **99.24%** | **99.24%** | **99.24%** |

Table 4: Comparison of Model Performances (**Dateset with background environment(noise)**)

| Model | Phylum Acc. | Class Acc. | Order Acc. | Family Acc. | Genus Acc. | Species Acc. |
|---|---|---|---|---|---|---|
| Efficient-Net [22] | 36.28% | 68.58% | 12.39% | 9.73% | 9.73% | 9.73% |
| DenseNet [11] | 96.02% | 98.23% | 94.69% | 95.13% | 95.13% | 95.13% |
| ResNet [10] | 39.28% | 68.58% | 9.73% | 9.73% | 9.73% | 9.73% |
| VGG [19] | 90.21% | 97.34% | 88.70% | 87.90% | 87.23% | 87.04% |
| VGG-MSSA | 91.15% | 96.90% | 90.71% | 91.15% | 91.15% | 91.15% |
| ResNet-MSSA | 46.46% | 69.91% | 27.88% | 23.89% | 35.84% | 26.55% |
| **Proposed Method** | **98.23%** | **99.12%** | **98.23%** | **97.35%** | **97.35%** | **98.20%** |

Table 3 and 4 shows the model comparison of our model with various models. We can observe that in Table 3 our proposed method outperforms all the models in all the taxonomic categories respectively. Our model achieved a classification accuracy of **99.24%** in all taxonomic division. DenseNet also performs well in all the respective taxonomic classification as compared to our proposed method. DenseNet reuses the feature maps, which helps it to perform better classification as compared to other models. When we look at the results obtained by different models on dataset which contains the background environment (Table 4), we can see that despite the background artifacts our model outperforms all the models again in all the taxonomic categories. The drop in classification accuracy in our model is very minimal as compared to other methods. The largest drop (classification accuracy) in our method was seen in Order Classification which was about **2.65%**. For the other categories the drop was not more than 2%.

If we compare the classification accuracies obtained by other model in the latter dataset, we can clearly see that accuracies have dropped significantly when there are background artifacts. For EfficientNet, phylum classification accuracy dropped from **76.63%** to **36.28**. For VGG it dropped from **96.24%** to **90.21%**. Similar trend can be observed for ResNet as well where phylum and class accuracy dropped from **78.63%** and **93.89%** to **39.29%** and **68.58%**.

In both the dataset integration of MSSA module to VGG and ResNet led to better classification accuracy across all taxonomic categories. Inclusion of MSSA improved their performance because it helped in capturing long-range dependencies in feature maps. Inclusion of MSSA not only enhanced classification but also reduced the model size and memory.

## 5.2 Unknown/New Species Classification Performance

In this section performance of various models on the task of classifying unknown species is compared in Tables 6 and 7, with and without background artifacts respectively. The results show that the proposed method consistently outperforms other models across multiple taxonomic levels. Four unknown species were considered for classification. They were Wood Sorel, Noni, Oleander and Jackfruit. Unknown species description is given in Table 5.

Table 5: Taxonomic classification of Wood Sorel, Noni, Oleander, and Jackfruit

| **Species Name** | **Phylum** | **Class** | **Order** | **Family** | **Genus** | **Species** |
|---|---|---|---|---|---|---|
| Wood Sorel | Magnoliophyta | Magnoliopsida | Oxalidales | Oxalidaceae | Oxalis | Oxalis acetosella |
| Noni | Magnoliophyta | Magnoliopsida | Gentianales | Rubiaceae | Morinda | Morinda citrifolia |
| Oleander | Magnoliophyta | Magnoliopsida | Gentianales | Apocynaceae | Nerium | Nerium oleander |
| Jackfruit | Magnoliophyta | Magnoliopsida | Rosales | Moraceae | Artocarpus | Artocarpus heterophyllus |

For images with background artifacts (Table 6), the proposed method achieves the highest Phylum Accuracy at **82.37%**, surpassing the next best model (VGG) by 1.14%. It also achieves the highest Class Accuracy at **76.39%**, which is a significant improvement over VGG's 72.21%. For Order Accuracy, the proposed method achieves **59.32%**, which is substantially higher than VGG's 46.29%.

It is evident from the results obtained that for images without background artifacts (Table 7), the proposed method again leads in Phylum Accuracy with **84.35%**, slightly greater than VGG. In terms of class accuracy VGG achieves the accuracy of **80.51%** where it is marginally ahead of the proposed method. The proposed method again achieves the highest Order Accuracy at **61.37%**, outperforming VGG's 57.23%. It also shows notable performance in Family Accuracy, achieving **43.32%**, where other models were not able to classify them. The superior performance of the proposed method is due to the integration of multi-scale self-attention module. This module helps the model focus on different parts of the image at various scales, which is particularly useful for recognizing complex patterns and details and retain taxonomic features. By capturing long range dependencies and important features, the model becomes more effective at distinguishing between different categories, even when the images have background artifacts. Additionally, the proposed method uses a larger receptive field, allowing it to gather more contextual information from the images. This leads to better feature representation and overall improved classification accuracy across various taxonomic levels.

Table 6: Classification Accuracy obtained on Unknown Species by different models (On images with background artifacts)

| **Model** | **Phylum Accuracy (%)** | **Class Accuracy (%)** | **Order Accuracy (%)** | **Family Accuracy (%)** |
|---|---|---|---|---|
| VGG | 81.23% | 72.21% | 46.29% | - |
| ResNet | 21.23% | 11.27% | - | - |
| Efficient-Net | 15.21% | - | - | - |
| Dense-Net21 | 79.21% | 61.12% | 34.71% | - |
| VGG-MSSA | 74.87% | 68.43% | 41.45% | - |
| ResNet50-MSSA | 18.61% | - | - | - |
| **Proposed Method** | **82.37%** | **76.39%** | **59.32%** | - |

Table 7: Classification Accuracy obtained on Unknown Species by different models (On images without background artifacts)

| **Model** | **Phylum Accuracy (%)** | **Class Accuracy (%)** | **Order Accuracy (%)** | **Family Accuracy (%)** |
|---|---|---|---|---|
| VGG | 84.23% | **80.51%** | 57.23% | - |
| ResNet | 28.23% | 18.23% | - | - |
| Efficient-Net | 11.21% | - | - | - |
| Dense-Net21 | 73.21% | 64.12% | 39.75% | - |
| VGG-MSSA | 81.87% | 77.43% | 46.45% | - |
| ResNet50-MSSA | 59.87% | 31.23% | - | - |
| **Proposed Method** | **84.35%** | 80.22% | **61.37%** | **43.32%** |

## 5.3 Cascaded Classifier Performance Comparison for various models

Moreover, we evaluated the performance of our cascaded classifier across all taxonomic categories for various models. Figures 7,9 and 11 shows the comparison of precision, recall and F1-scores achieved by various model on the dataset which had no background-artifacts. Similarly Figures 8,10 and 12 shows the comparison of precision, recall and F1-scores achieved by various model on the dataset which had background artifacts such as external objects, light etc. In the grouped histogram plots each colour represents a model. To better understand the legend for histogram plot is shown in Figure 6.

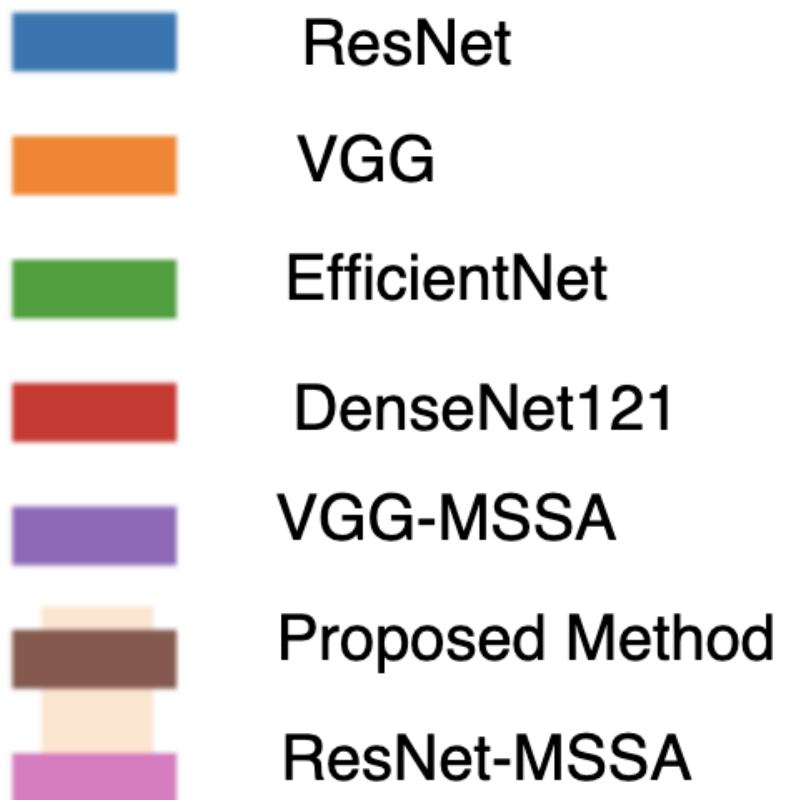

Figure 6: Legend for histogram plots

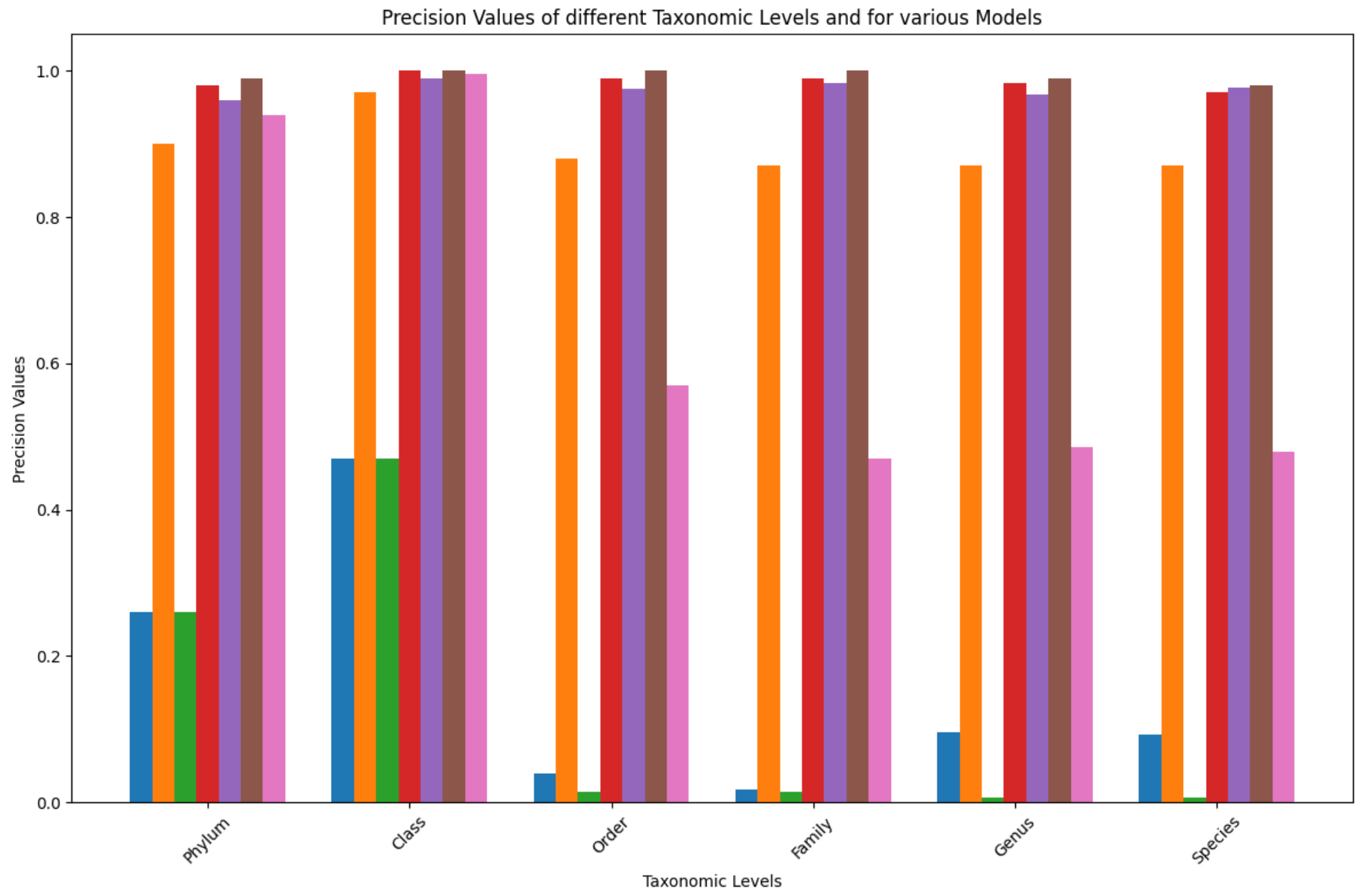

Figure 7: Precision values for various models **(dataset without background artifacts)**

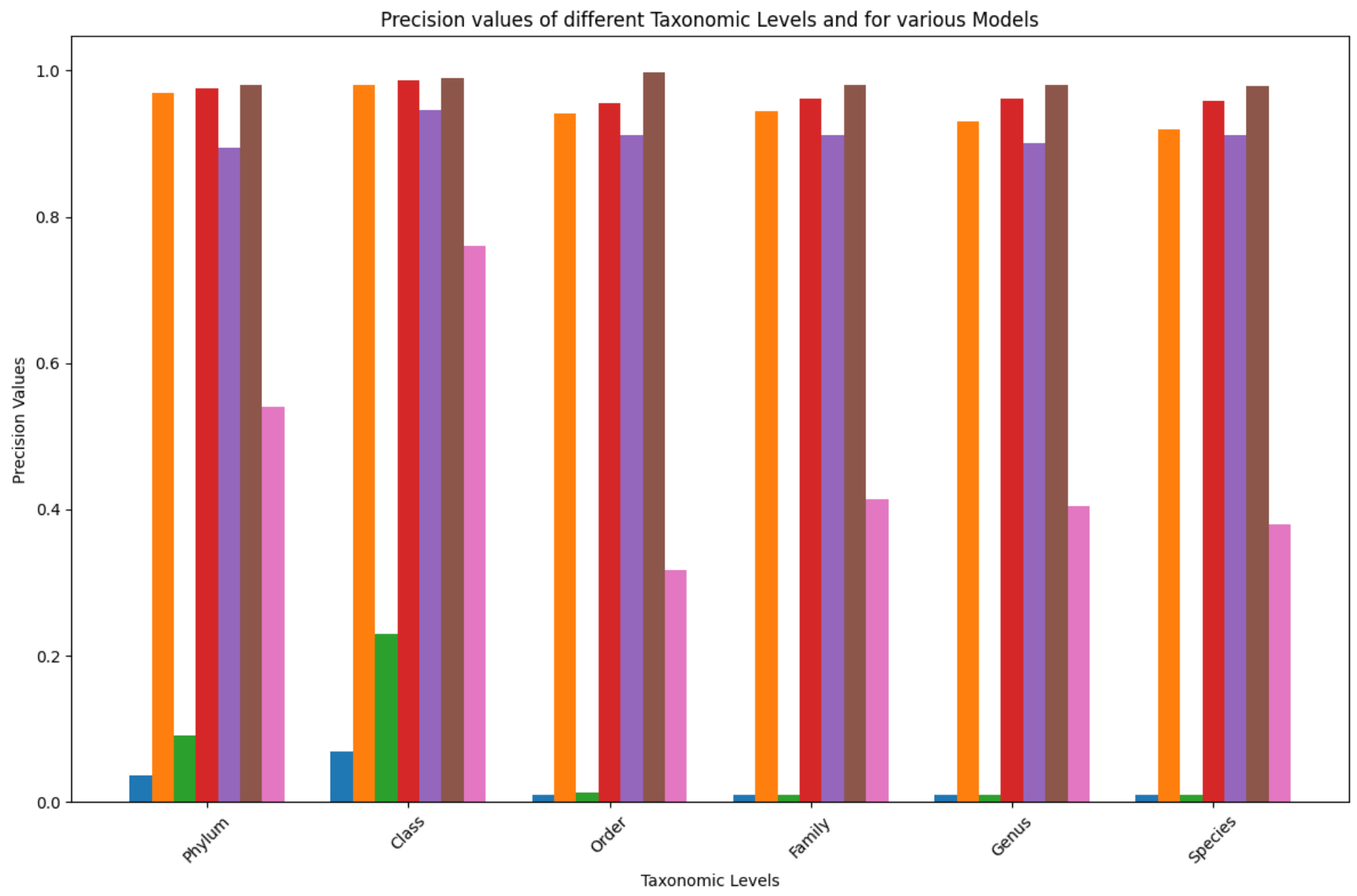

Figure 8: Precision values for various models **(dataset with background artifacts)**

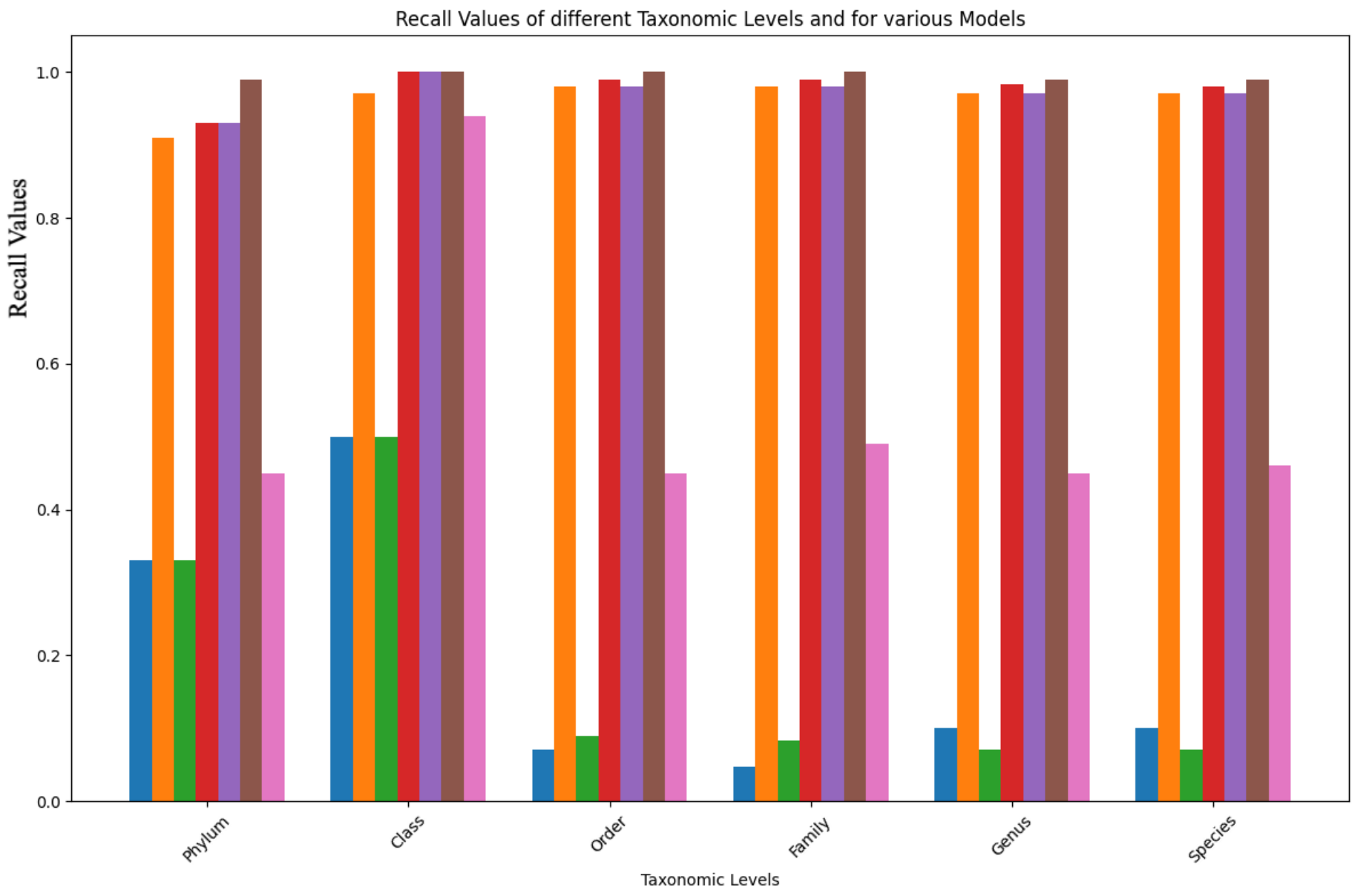

Figure 9: Recall values for various models **(dataset without background artifacts)**

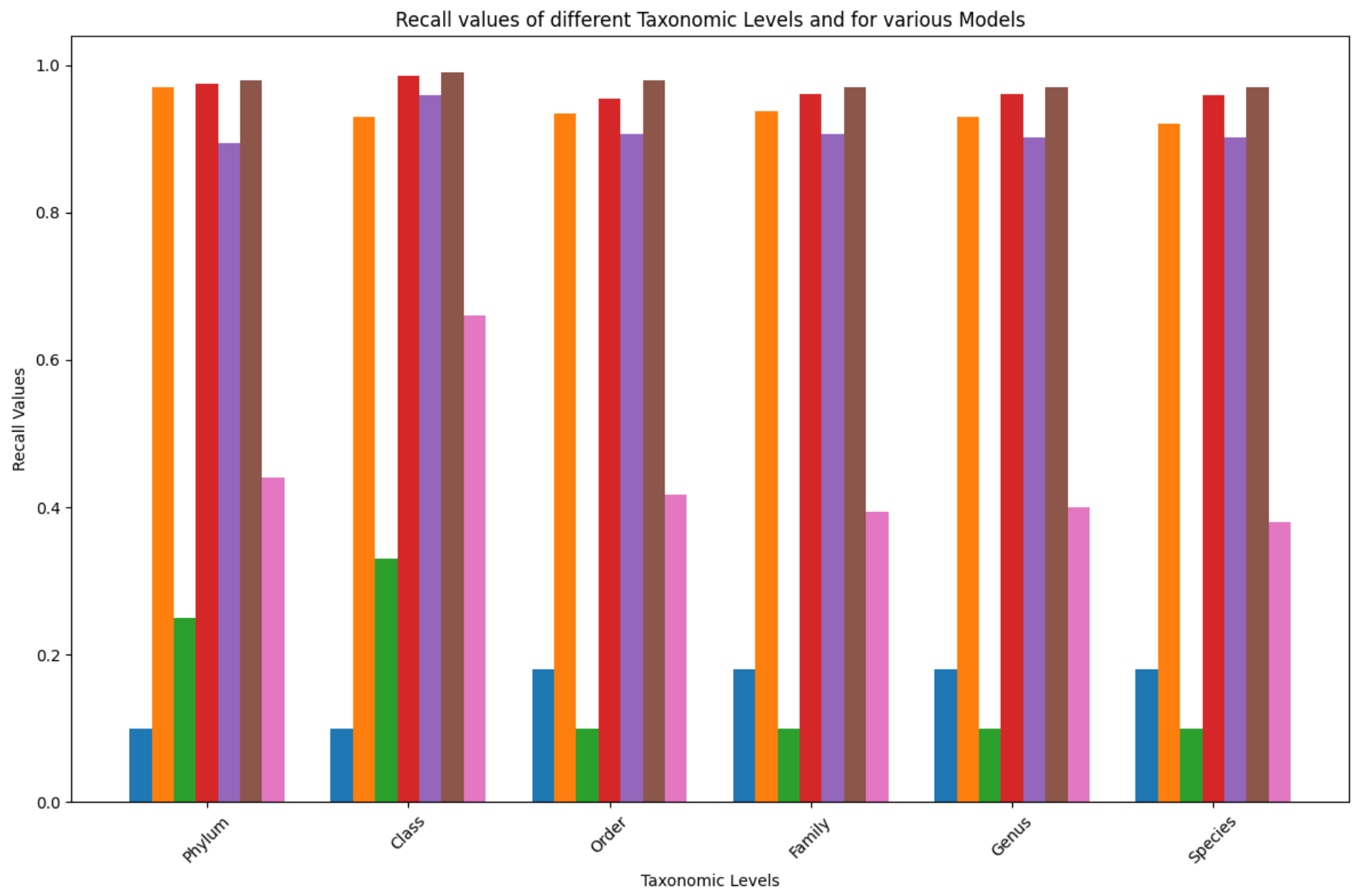

Figure 10: Recall values for various models **(dataset with background artifacts)**

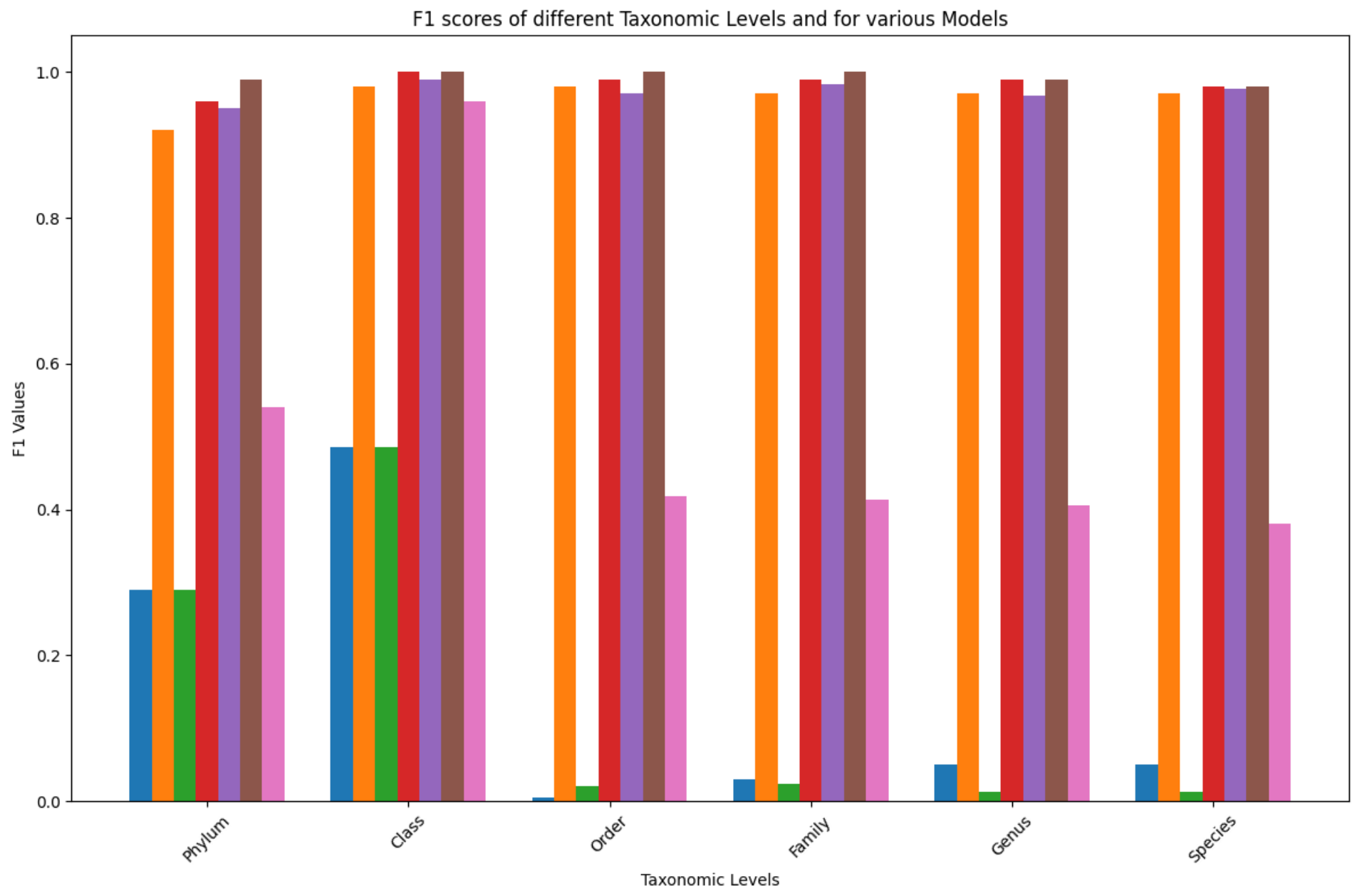

Figure 11: F1-scores for various models **(dataset without background artifacts)**

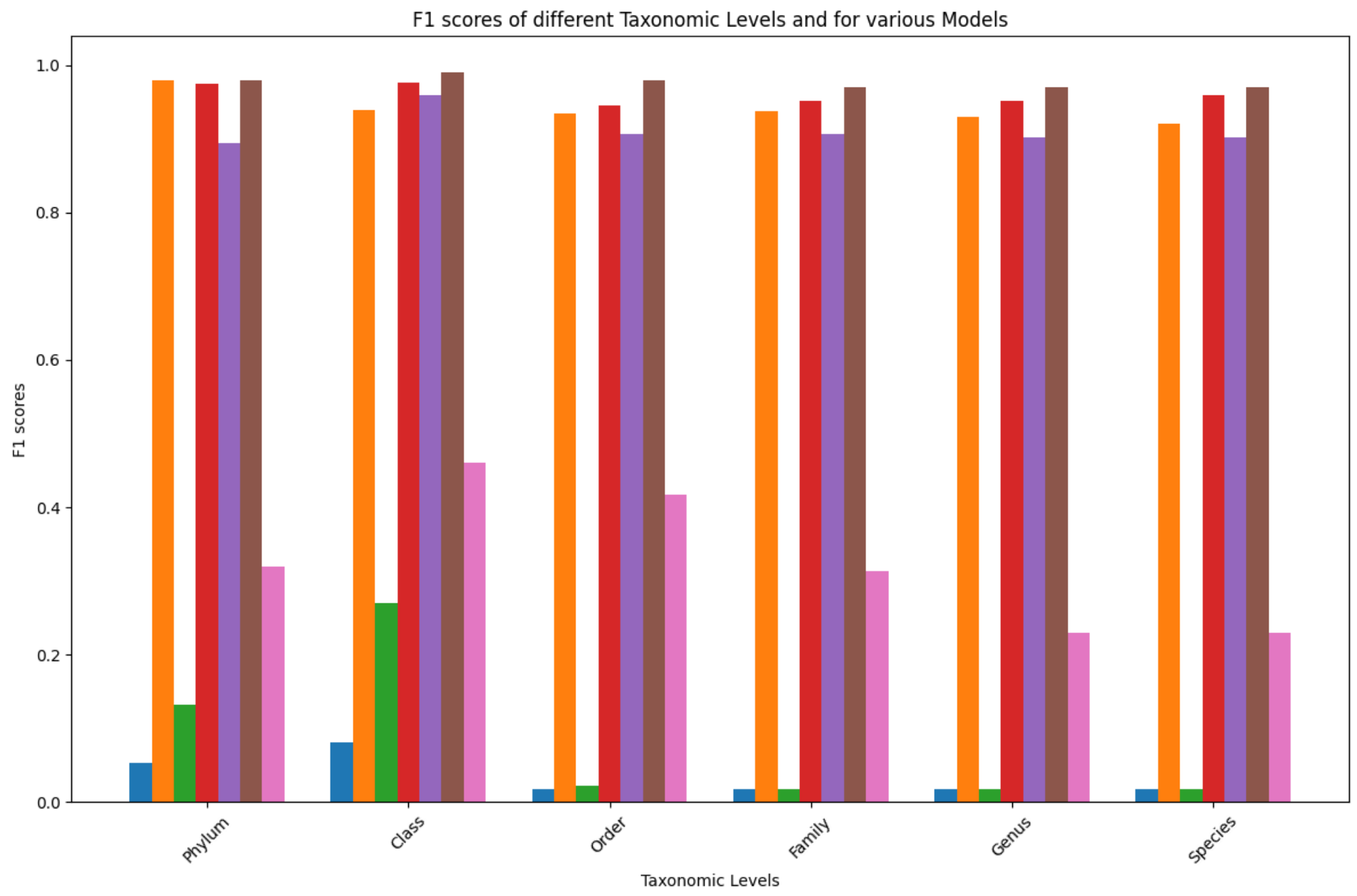

Figure 12: F1-scores for various models **(dataset with background artifacts)**

## 5.4 Ablation Study

In this ablation study, we evaluate the impact of combining different techniques on model performance across various taxonomic levels. We consider six combinations of models: DenseNet , DenseNet with MSSA, ResNet, ResNet with MSSA, VGG, and VGG with MSSA. We report the precision/recall and F1 score for each combination on both the old dataset (with background artifacts) and the new dataset (without background artifacts) in Table 8-11. The text in bold represents the test classification accuracy obtained by the proposed architecture

Table 8: Precision/Recall on Dataset with background artifacts

| **Model** | **Phylum** | **Class** | **Order** | **Family** | **Genus** | **Species** |
|---|---|---|---|---|---|---|
| DenseNet + Cascaded Classifier | 0.97/0.97 | .98/.98 | 0.95/0.95 | 0.96/0.95 | 0.96/0.96 | 0.95/0.95 |
| **DenseNet + MSSA + Cascaded Classifier** | **0.98/.98** | **.99/.99** | **0.99/.98** | **.98/.97** | **0.98/.97** | **0.97/.97** |
| ResNet + Cascaded Classifier | 0.036/0.1 | 0.069/0.1 | 0.010/0.18 | 0.010/0.18 | 0.01/0.18 | 0.01/0.18 |
| ResNet + MSSA + Cascaded Classifier | 0.54/0.44 | 0.76/0.66 | 0.31/0.41 | 0.41/0.39 | 0.40/0.40 | 0.38/0.38 |
| VGG + Cascaded Classifier | 0.97/0.97 | 0.98/0.93 | 0.94/0.93 | 0.94/0.93 | 0.93/0.93 | 0.92/0.92 |
| VGG + MSSA + Cascaded Classifier | 0.89/0.89 | 0.94/0.93 | 0.91/0.90 | 0.91/0.90 | 0.90/0.90 | 0.91/0.90 |

Table 9: F1 Score on Old Dataset with background artifacts

| **Model** | **Phylum** | **Class** | **Order** | **Family** | **Genus** | **Species** |
|---|---|---|---|---|---|---|
| DenseNet + Cascaded Classifier | 0.975 | .976 | 0.945 | 0.951 | 0.951 | 0.959 |
| **DenseNet + MSSA + Cascaded Classifier** | **0.98** | **.99** | **.98** | **.97** | **.97** | **.97** |
| ResNet + Cascaded Classifier | 0.053 | 0.081 | 0.018 | 0.018 | 0.018 | 0.018 |
| ResNet + MSSA + Cascaded Classifier | 0.32 | 0.46 | 0.417 | 0.314 | 0.23 | 0.23 |
| VGG + Cascaded Classifier | **0.98** | 0.94 | 0.934 | 0.938 | 0.93 | 0.92 |
| VGG + MSSA + Cascaded Classifier | 0.895 | 0.96 | 0.906 | 0.90 | 0.90 | 0.902 |

Table 10: Precision/Recall on New Dataset without background artifacts

| **Model** | **Phylum** | **Class** | **Order** | **Family** | **Genus** | **Species** |
|---|---|---|---|---|---|---|
| DenseNet + Cascaded Classifier | 0.996/0.93 | **1.00/1.00** | 0.99/0.99 | 0.99/0.99 | 0.99/0.98 | 0.98/0.98 |
| **DenseNet + MSSA + Cascaded Classifier** | **0.99/0.99** | **1.00/1.00** | **1.00/1.00** | **1.00/1.00** | **0.99/0.99** | **0.98/0.99** |
| ResNet + Cascaded Classifier | 0.26/0.33 | 0.47/0.50 | 0.04/0.07 | 0.017/0.0475 | 0.095/0.10 | 0.092/0.10 |
| ResNet + MSSA + Cascaded Classifier | 0.94/0.45 | 0.995/0.94 | 0.57/0.45 | 0.47/0.49 | 0.485/0.45 | 0.48/0.46 |
| VGG + Cascaded Classifier | 0.90/0.91 | 0.97/0.97 | 0.88/0.98 | 0.87/0.98 | 0.87/0.97 | 0.87/0.97 |
| VGG + MSSA + Cascaded Classifier | 0.96/0.93 | 0.99/1 | 0.976/0.98 | 0.983/0.98 | 0.967/0.97 | 0.977/0.97 |

Table 11: F1 Score on New Dataset without background artifacts

| **Model** | **Phylum** | **Class** | **Order** | **Family** | **Genus** | **Species** |
|---|---|---|---|---|---|---|
| DenseNet + Cascaded Classifier | 0.96 | **1.00** | 0.99 | 0.99 | 0.98 | 0.97 |
| **DenseNet + MSSA + Cascaded Classifier** | **0.99** | **1.00** | **1.00** | **1.00** | **.99** | **0.98** |
| ResNet + Cascaded Classifier | 0.29 | 0.485 | 0.005 | 0.03 | 0.05 | 0.05 |
| ResNet + MSSA + Cascaded Classifier | 0.54 | 0.96 | 0.4175 | 0.414 | 0.405 | 0.38 |
| VGG + Cascaded Classifier | 0.92 | 0.98 | 0.98 | 0.97 | 0.97 | 0.97 |
| VGG + MSSA + Cascaded Classifier | 0.95 | 0.99 | 0.97 | 0.9833 | 0.967 | 0.977 |

#### 5.4.1 Discussion

The integration of MSSA module enhances performance by integrating local spatial details with high-level semantic information from different scales. Using attention scores, MSSA focuses on important features across multiple scales, helping the model capture complex relationships in the data, leading to higher precision and recall, as seen in the improved metrics for all the taxonomic categories. By considering features at multiple scales, MSSA allows the model to understand both fine-grained and broad patterns in the images, which is particularly useful for distinguishing between closely related species, resulting in higher accuracy for Order and Species.

We extracted the feature maps from the pooling layers of the model and from the final layer of the MSSA module.

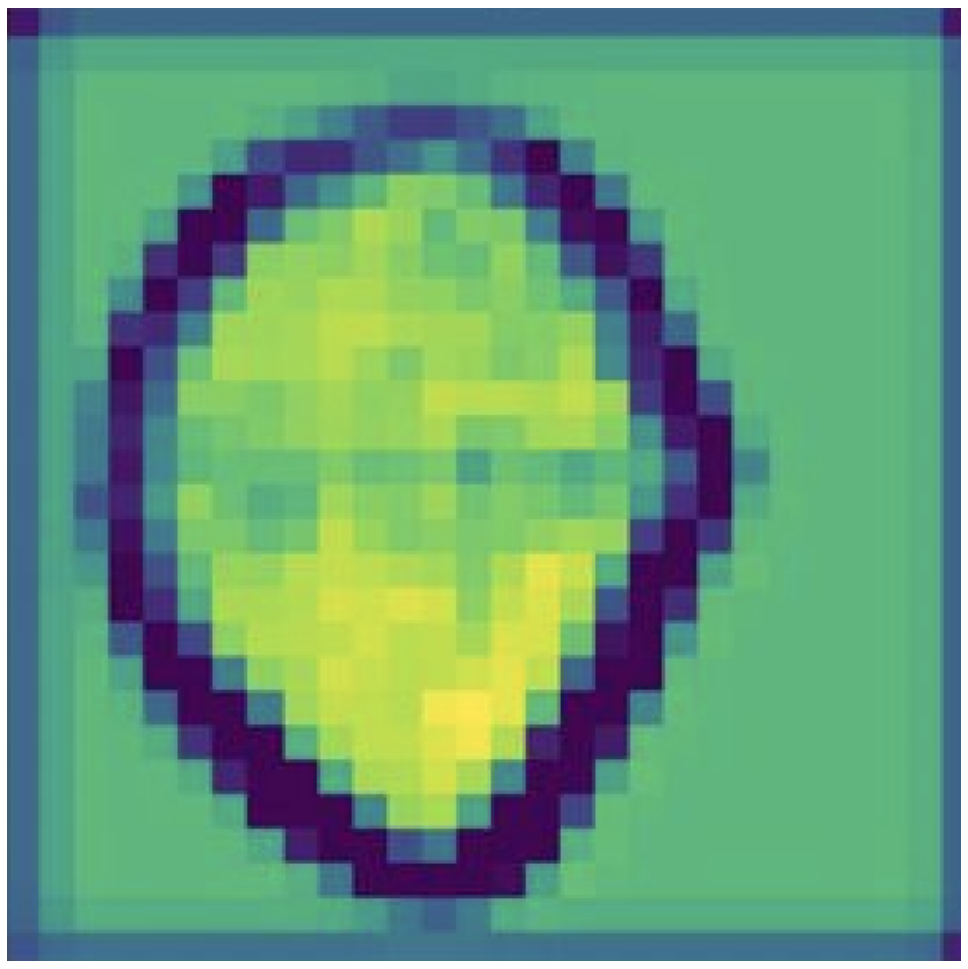

(a) Feature map extracted from 2nd pooling layer of DenseNet

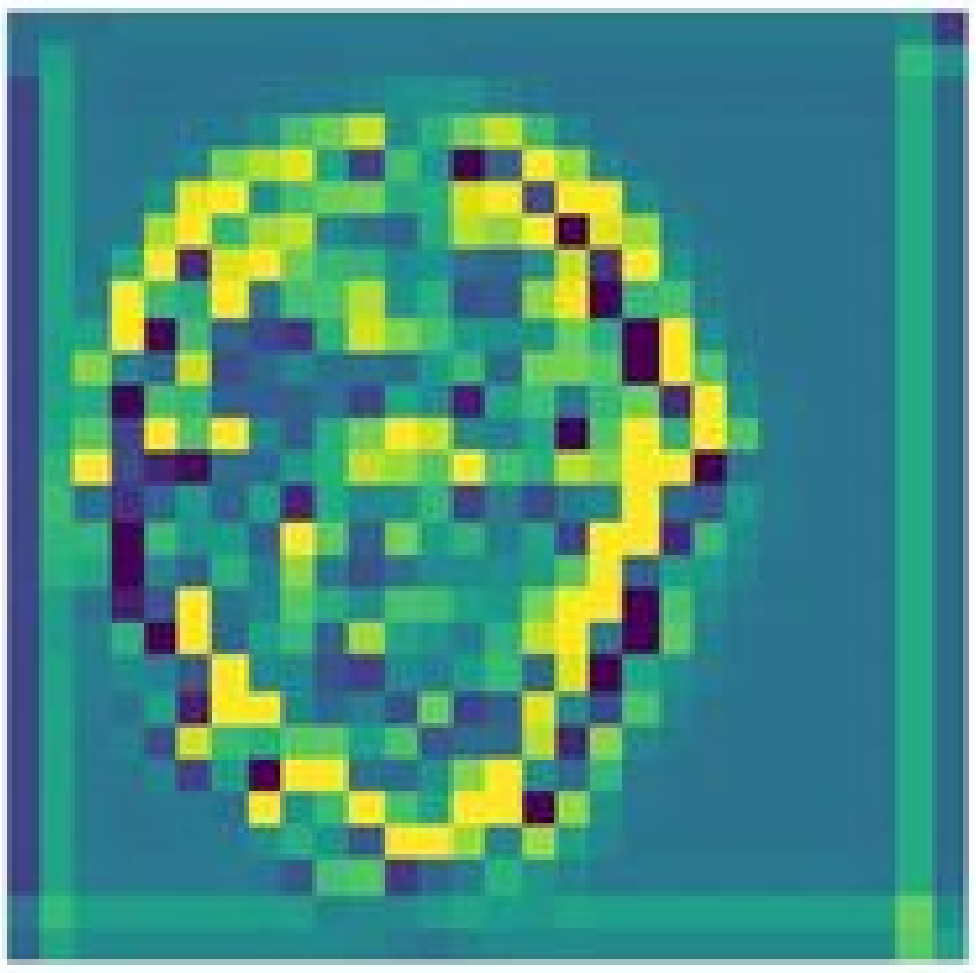

(b) Feature map extracted from final layer of MSSA module

Figure 13: Dataset without background artifacts

The first image (13a) shows the feature map extracted from the second pooling layer of DenseNet. This feature map predominantly captures low-level features, which include basic spatial details and textures. However, it lacks the ability to capture long-range dependencies and high-level semantic information. The spatial details are somewhat clear, but the overall representation is not sufficient for complex tasks such as distinguishing between similar species or identifying new species. The second image (13b) shows the feature map from the final layer of the MSSA module. Compared to the feature map from the pooling layer, this map is more refined and detailed. MSSA enhances feature representation by combining both local spatial details and high-level semantic information. This is evident in the clarity and distinctiveness of the features in the map.

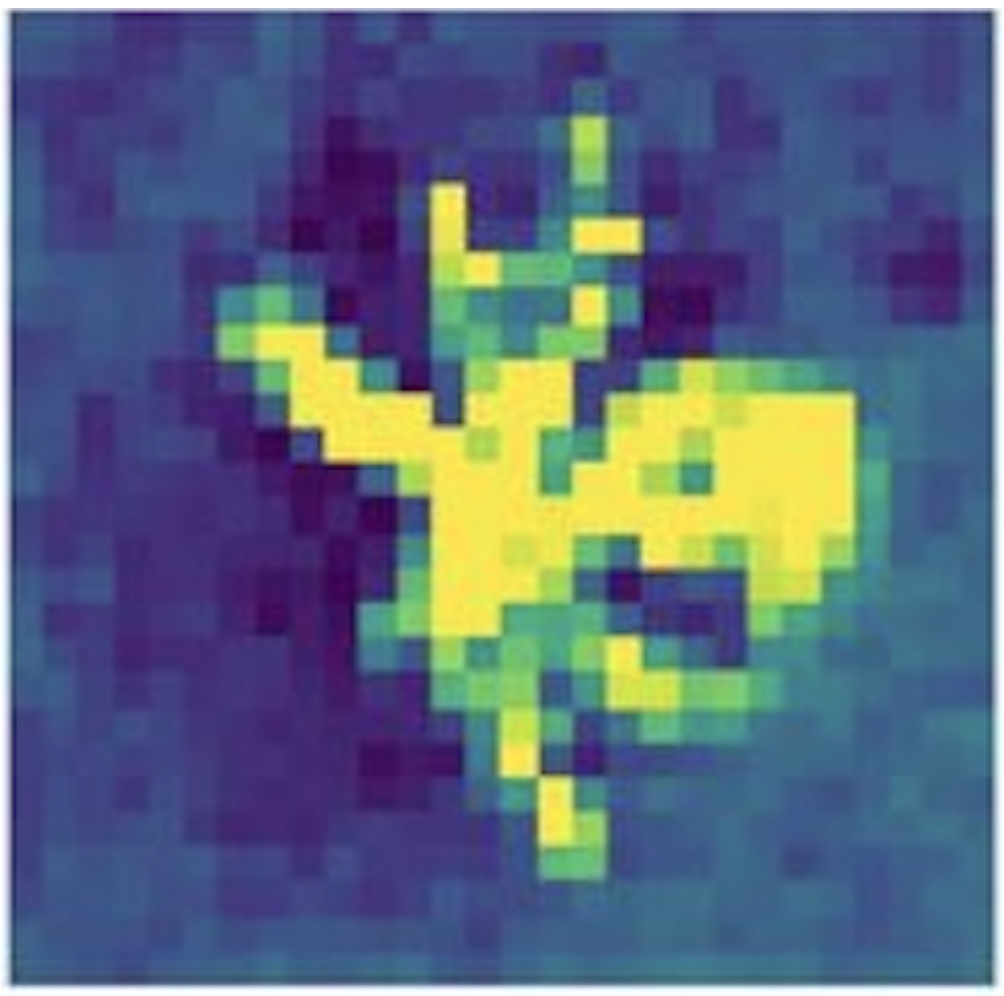

(a) Feature map extracted from 2nd pooling layer of DenseNet

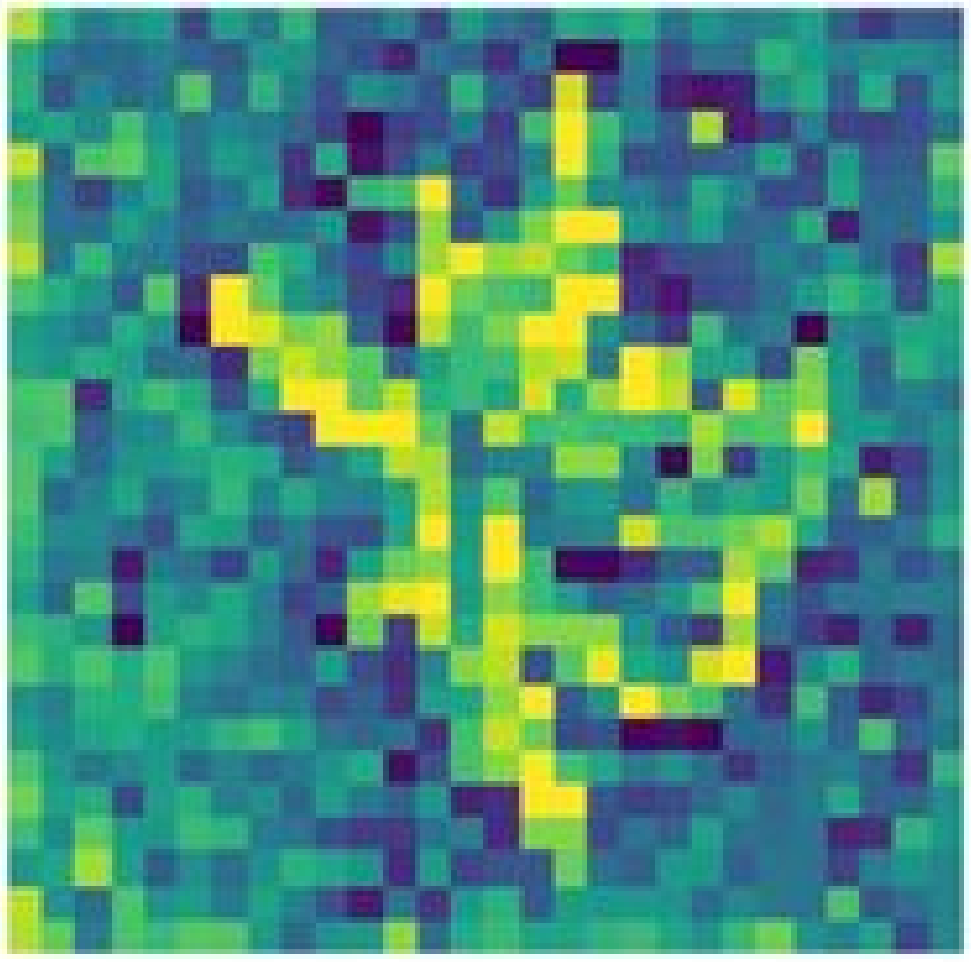

(b) Feature map extracted from final layer of MSSA module

Figure 14: Dataset without background artifacts

Similarly feature map is extracted from the second pooling layer of DenseNet(noisy data) and is shown in Fig (14a). This feature map primarily captures the local spatial details however it lacks high-level semantic information. The feature map shows a basic representation with clear boundaries but does not capture the intricate contextual. The second image (14b) captures the feature map from the final layer of the MSSA module on the noisy dataset. In the old dataset, background artifacts can obscure the key characteristics of the medicinal plants, leading to misclassifications. However the model performs better in cleaner dataset as compared to the latter one. Despite the images affected with noise in old dataset we are able to acheive good results because of the MSSA module which helps to focus on important features by supressing the noise.

Furthermore, addition of MSSA improves robustness to noisy data by giving importance relevant features and ignoring unnecessary background artifacts. This fact can be checked from the performance drop seen on the dataset with background artifacts when trained on models without MSSA. Additionally, MSSA enhances the model's ability to understand the context of features within an image, which is crucial for hierarchical classification tasks where understanding the relationship between different taxonomic levels is important.

### 5.4.2 Impact of MSSA on Performance

Tables 8, 9, 10, and 11 show the precision, recall, and F1-score metrics for different models on both datasets. The results clearly demonstrate that incorporating MSSA significantly improves the performance across all taxonomic levels.

For the dataset with background artifacts, the DenseNet + MSSA achieves a precision and recall of 0.98 for Phylum, 0.99 for Class, 0.99 for Order, 0.98 for Family, 0.98 for Genus, and 0.97 for Species. These values are higher than those of the DenseNet without MSSA, which shows the effectiveness of MSSA in capturing important features within the data.

Similarly, for the dataset without background artifacts, DenseNet + MSSA achieves fairly good precision and recall across all taxonomic levels. For example, it achieves 0.99 for Phylum, 1.00 for Class, and 0.98 for Species. These results demonstrate that the model performs even better with clean data.

The F1-scores further highlight the benefits of MSSA. For the dataset with background artifacts, DenseNet + MSSA achieves F1-scores of 0.98 for Phylum, 0.99 for Class, 0.98 for Order, and 0.97 for Species. These scores represent an improvement over models without MSSA, indicating better overall classification performance. On the dataset without background artifacts, it achieves F1-scores of 0.99 for Phylum, 1.00 for Class, and 0.98 for Species, again outperforming other models.

## 5.5 Model Size and Parameters

In this section we present a comparison of our proposed approach with respect to other architectures on model size and parameters. As shown in Table 12, our model consists of **6,360,879** parameters, which translates to a memory size of just **24.26 MB**. In contrast, the DenseNet21 model has 33,557,116 parameters and requires 128.01 MB of memory. Similarly, the EfficientNet model, with 36,991,711 parameters, occupies 141.11 MB. Even more substantial models like ResNet100, with 75,797,436 parameters, demand 289.14 MB of memory. This significant reduction in model size and memory footprint offers several advantages. Firstly, it makes the model highly suitable for deployment on resource constrained devices, such as mobile phones and embedded systems, which are commonly used in field research and conservation efforts. The compactness of our model ensures that it can be utilized in real-time scenarios without incurring substantial computational overhead, making it practical for on-the-go applications.

Furthermore, the reduced memory requirement enhances the model's scalability and responsiveness, allowing it to process and classify medicinal plant images more swiftly and efficiently. This efficiency does not come at the cost of accuracy, as our model maintains good performance metrics in both the datasets. This presents our model as an optimal balance between effectiveness and resource efficiency, making it an ideal solution for real-world applications in medicinal plant classification.

Table 12: Number of Parameters and Memory Size

| **Model** | **Number of Parameters** | **Memory Size (MB)** |
|---|---|---|
| VGG [19] | 28389244 | 108.30 (MB) |
| ResNet100 [10] | 75797436 | 289.14 (MB) |
| Efficient-Net [22] | 36991711 | 141.11 (MB) |
| Dense-Net21 [11] | 33557116 | 128.01 (MB) |
| VGG-MSSA | 18158164 | 69.27 (MB) |
| ResNet-MSSA | 15164692 | 57.85 (MB) |
| **Proposed Method** | **6360879** | **24.26 (MB)** |

# 6 Conclusion

In this study, we introduced a new model which integrates DenseNet121 with a Multi-Scale Self-Attention (MSSA) mechanism and a cascaded classifier for the hierarchical classification of medicinal plants. Our experiments on datasets with and without background artifacts demonstrated that the inclusion of MSSA significantly enhances the model's performance, achieving higher precision, recall, and F1-scores across all taxonomic levels compared to traditional classifiers. The MSSA module is beneficial because it incorporates attention scores to capture complex relationships within the data, integrating local spatial details with high-level semantic information from different scales. This capability allows the model to handle noisy data and accurately distinguish between closely related species, making it well-suited for real world applications where image data may be degraded due to environmental factors. Our model outperformed other models in our study, particularly while handling dataset with background artifacts, demonstrating its effectiveness. Our results highlighted the importance of incorporating MSSA in hierarchical classification tasks, paving the way for more accurate and reliable medicinal plant classification. Our model when tested on unknown species provided promising results. By accurately predicting higher taxonomic levels such as phylum, class, and order, the model can generalize and classify unknown species that share these characteristics with known species. This hierarchical approach ensures that even when encountering novel species, the model can still provide meaningful classifications based on shared taxonomic features, enhancing its utility in biodiversity research and conservation efforts. Overall, this model represents a significant advancement in the field of medicinal plant classification, providing a powerful tool for researchers and practitioners in botany, pharmacology, and related fields. Future work will explore further enhancements to the model, including the integration of additional attention mechanisms and the application to other hierarchical classification tasks.

**Competing Interests:** There are no competing interests.

**Funding Information**: Not Applicable

**Author Contribution**: All the authors have contributed equally.

**Data Availability Statement**: Dataset availability on request.

**Research Involving Human and /or Animals**: Not Applicable

**Informed Consent**: Not Applicable